\documentclass[10pt,twocolumn,letterpaper]{article}

\usepackage{iccv}
\usepackage{times}
\usepackage{algorithm}
\usepackage{algorithmic}
\usepackage{epsfig}
\usepackage{graphicx}
\usepackage{amsmath}
\usepackage{amssymb}
\usepackage{bm}
\usepackage{booktabs}
\usepackage{multirow}
\usepackage[cmyk]{xcolor}
\usepackage{utfsym}
\usepackage{subcaption}

\usepackage{colortbl}

\usepackage[pagebackref=true,breaklinks=true,letterpaper=true,colorlinks,bookmarks=false]{hyperref}

\iccvfinalcopy 

\def\iccvPaperID{7487} 

\ificcvfinal\fi

\begin{document}

\title{LafitE: Latent Diffusion Model with Feature Editing for \\Unsupervised Multi-class Anomaly Detection}

\author{
Haonan Yin $^{1, \star }$  \quad Guanlong Jiao $^{1, \star }$ \quad Qianhui Wu $^{2 }$ \quad Börje F. Karlsson $^{2 }$ \\ \quad Biqing Huang $^{1, \dagger }$ \quad Chin-Yew Lin $^{2 }$\\
$^1$Tsinghua University \quad $^2$Microsoft Research Asia\\
{\tt\small {yhn21, jgl22}@mails.tsinghua.edu.cn} \quad {{\tt\small hbq@tsinghua.edu.cn}}\\
{\tt\small {qianhuiwu ,borjekar, cyl}@microsoft.com} \\
}

\maketitle
\ificcvfinal\thispagestyle{empty}\fi

\begin{abstract}
In the context of flexible manufacturing systems that are required to produce different types and quantities of products with minimal reconfiguration, this paper addresses the problem of unsupervised multi-class anomaly detection:
develop a unified model to detect anomalies from objects belonging to multiple classes when only normal data is accessible.
We first explore the generative-based approach and investigate latent diffusion models for reconstruction to mitigate the notorious ``identity shortcut'' issue
in auto-encoder based methods.
We then introduce a feature editing strategy that modifies the input feature space of the diffusion model to further alleviate ``identity shortcuts'' and meanwhile improve the reconstruction quality of normal regions, leading to fewer false positive predictions.
Moreover, we are the first who pose the problem of hyperparameter selection in unsupervised anomaly detection, and propose a solution of synthesizing anomaly data for a pseudo validation set to address this problem.
Extensive experiments on benchmark datasets MVTec-AD and MPDD show that the proposed \textbf{LafitE}, \ie, \textbf{La}tent Di\textbf{f}us\textbf{i}on Model with Fea\textbf{t}ure \textbf{E}diting, outperforms state-of-art methods by a significant margin in terms of average AUROC. 
The hyperparamters selected via our pseudo validation set are well-matched to the real test set. 
\end{abstract}

\footnotetext{$^\star$ Equal Contribution Authors \quad $^\dagger$ Corresponding Authors}

\section{Introduction}
\label{sec:intro}
Anomaly detection (AD) is the task of identifying and localizing anomalous patterns that are atypical of those seen in normal instances.
It has attracted considerable attention from the research community in recent years for various application scenarios, such as industrial defect detection \cite{MVTec}, medical image analysis \cite{MedicalADSurvey}, and video inspection \cite{VideoADSurvey}.
However, there are significant difficulties in getting access to a large number of anomalous data samples because the types of anomalies usually show great diversity in the real world, ranging from texture level (\eg, scratches and holes) to semantic level (\eg, missing components).
Therefore, many works have been proposed for the extreme scenario, \ie, the \textit{unsupervised} setting, where no prior information on anomalies is available and models need to be developed utilizing normal data only \cite{US,DRAEM,PatchSVDD,PatchCore,RD4AD,UniAD,PaDiM}.

\begin{figure}[t]
	\centering
	\includegraphics[width=0.99\linewidth]{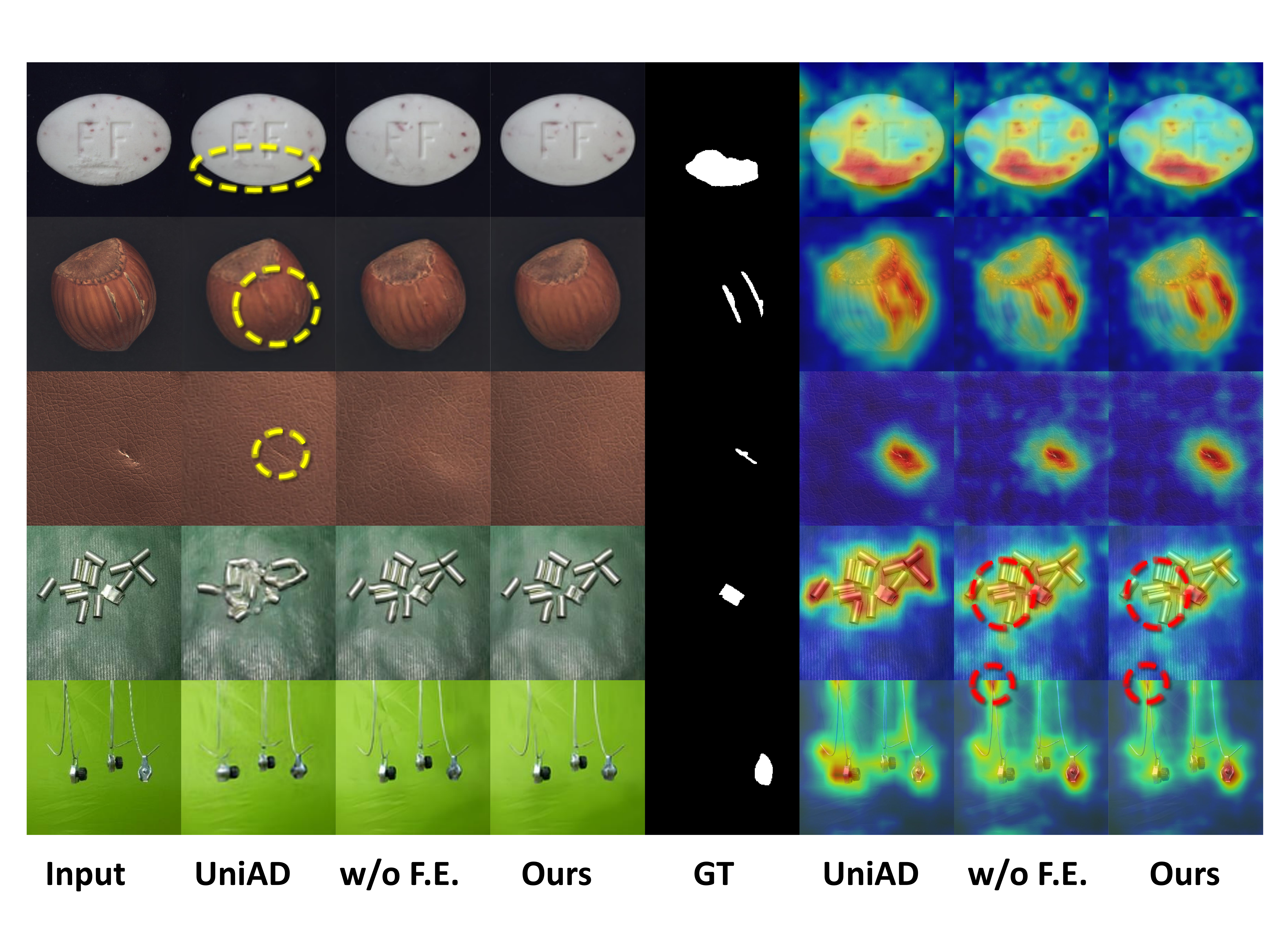}
        \hfill
	\caption{Visualization of the reconstructed images and the predicted heatmap of anomalous regions obtained from the baseline model UniAD and our approach.
    ``GT'': the ground-true anomalous region.
    ``w/o FE'': our approach without feature editing.}
	\label{fig:motivation}
\end{figure}

Flexible manufacturing systems are renowned for their ease of adapting to changes in product specifications and quantities,
and are gaining increasing popularity for various benefits such as cost reduction, increased productivity, and shortened lead times.
In such a context involving objects from multiple classes, most prior studies on unsupervised anomaly detection suggest training separate models for different classes \cite{regad,metaformer}, which could be memory-intensive and rigid.

Recently, UniAD \cite{UniAD} proposed a unified model for unsupervised multi-class anomaly detection.
It follows the \textbf{reconstruction} paradigm \cite{DRAEM,RIAD,metaformer} and utilizes the Transformer architecture \cite{vaswani2017attention} as the backbone of the auto-encoder.
The idea is that the auto-encoder trained on normal samples will produce low reconstruction error in normal regions but high reconstruction error in anomalous regions, hence allowing anomalies to be localized by the difference between an input sample and a reconstructed one.
UniAD proposes three strategies to mitigate the ``identity shortcut'' issue
, where the network may directly copy the input as the output regardless of whether it is normal or not, thus failing to localize the anomalies, \ie, leading to false negative predictions.
Though alleviating the problem, we can still observe the ``identity shortcut'' issue by visualizing their outputs as highlighted by yellow circles in Figure \ref{fig:motivation}.
We suggest the intrinsic nature of auto-encoder based methods, \ie, aiming to learn a mapping function to the normal data manifold with no guarantee for anomalies to be mapped as well \cite{DRAEM}, makes it prone to errors.

\begin{figure}[t]
	\centering
	\includegraphics[width=0.8\linewidth]{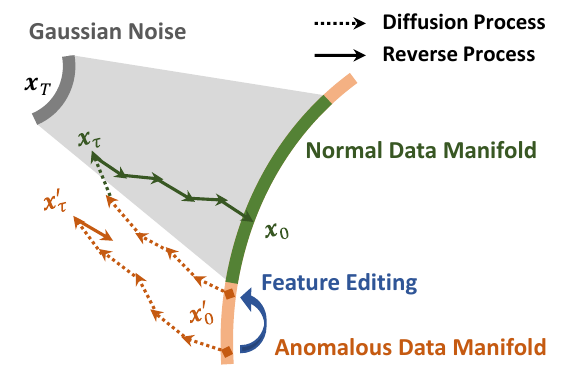}
	\caption{Illustration of how diffusion model and feature editing benefit the reconstruction. The diffusion process makes the normal and anomalous data less distinguishable in the near-Gaussian manifold, while feature editing further pulls the anomalous data closer to the normal data manifold in the input feature space.}
	\label{fig:motivation_FE}
\end{figure}

To remedy this, we study the use of generative models for performing reconstruction while better learning the multi-class normal data distribution in the latent space.
Diffusion models \cite{DDPM,DDIM,LDM} have recently achieved great success in image generation with high guarantees of density estimation \cite{VDM} and sampling quality \cite{DMbeatsGAN}.
We argue that using the diffusion model for anomaly reconstruction is a better paradigm in nature:
The diffusion model learns the process from a Gaussian distribution to the normal distribution;
The anomalous samples would be quite close to the normal ones on the near Gaussian manifold as illustrated in Figure \ref{fig:motivation_FE} and Figure \ref{fig:latent_vis}.
In this way, they will have a higher probability to be reconstructed to normal samples via the denoising generation process of the diffusion model.
Therefore, in this work, we take the diffusion model as the backbone of our proposed approach.
We propose \textbf{LafitE}, \ie, \textbf{La}tent Di\textbf{f}fus\textbf{i}on Model with Fea\textbf{t}ure \textbf{E}diting, to address unsupervised multi-class anomaly detection.
Specifically, we first map input images into a latent feature space: We extract hierarchical patch representations using a model pre-trained on ImageNet, and produce a feature tensor for each image by aggregating multi-scale representations.
With such aggregation, the feature slice corresponding to each position in the raw image contains both low-level texture information and high-level semantic information, which allows the method to deal with anomalies of various dimensions.
Then, we employ a diffusion model \cite{DDIM} to reconstruct the aggregated feature tensor in the latent space.
During inference, we first run the diffusion process to a  maximum corruption step $\tau$, and then perform the denoising process from the corrupted feature tensor for reconstruction.
In view of the fact that patch representations containing anomalous information may lead to corrupted feature tensors far from the normal data manifold (\eg, $x'_{\tau}$ in Figure \ref{fig:motivation_FE}), we further propose a feature editing scheme over the \textit{input space} of the diffusion model to pull patch representations corresponding to anomalies much closer to the normal data manifold.
Particularly, we use a memory bank to store typical patch representations of seen normal data.
During inference, we replace each patch representation of the feature tensor with its weighted nearest neighbors in the memory bank and perform reconstruction afterward with the learned diffusion model.
Such feature editing brings benefits from two perspectives:
i) Anomalous patch representations are directly dropped at the input stage of the reconstruction model and therefore the ``identity shortcut'' issue can be naturally mitigated.
ii) The edited feature tensors locate
much closer to the normal data manifold, which facilitates better reconstruction performance and thus reduces false positive predictions as highlighted by red circles in Figure \ref{fig:motivation}.


Due to the inherent characteristics of unsupervised AD that only normal samples are available for learning, how to select hyperparameters for model training and inference becomes a critical issue but has not been well discussed in previous work.
In this paper, we first pose the problem and further propose a simple but effective method to address it: We construct a pseudo validation set by leveraging the normal samples to synthesize anomalous samples.
Experiments results indicate that the model's performance on our pseudo validation set is generally consistent with that on the real test.
The results also show that our LafitE significantly outperforms prior methods with a large margin in terms of average AUROC on both MVTec-AD and MPDD.

\section{Related Work}

\begin{figure*}[t]
\centering
\includegraphics[width=0.9\linewidth]{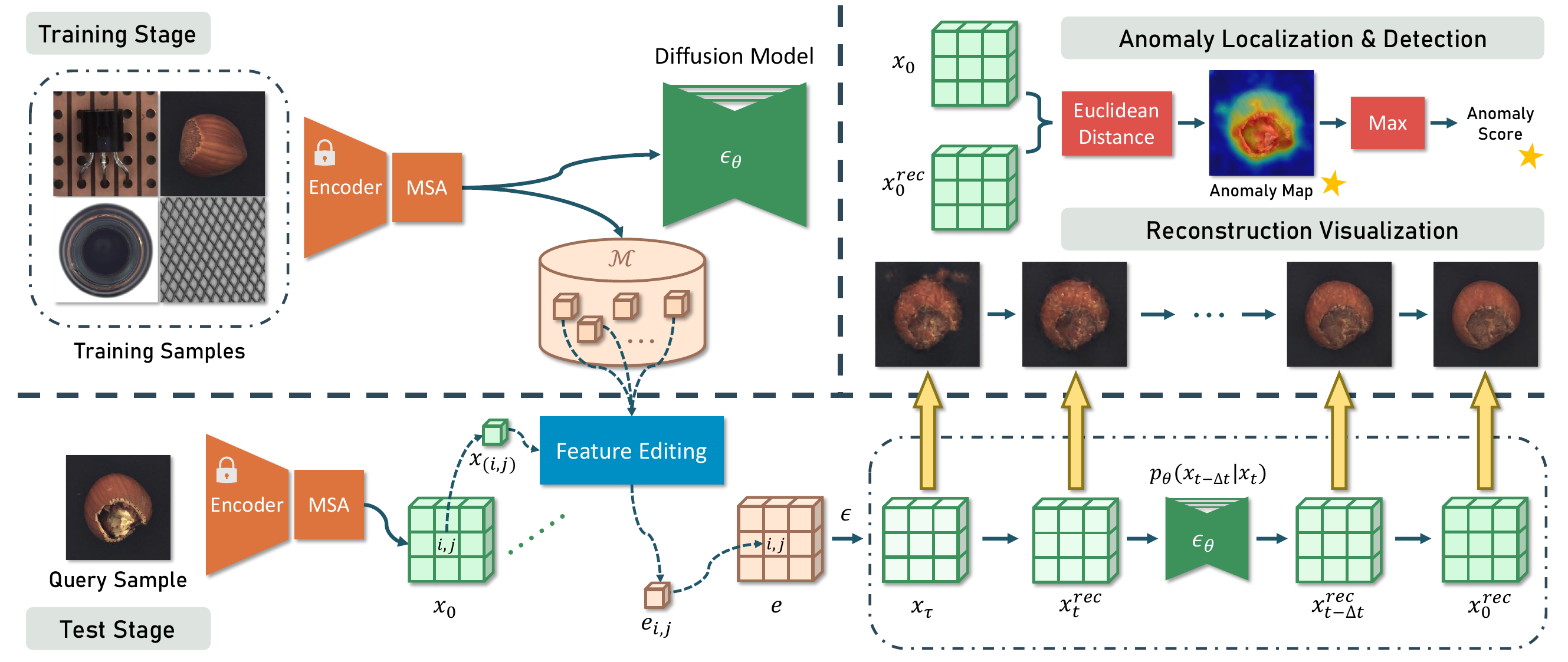}
	\caption{Overall framework of the proposed LafitE: It learns a diffusion model capable of modeling normal distribution during the training stage and stores a memory bank for feature editing. During the test stage, features of the query sample are first edited to ones closer to the normal data manifold. Then, the edited features are iteratively reconstructed via the denoising process of the diffusion model. Finally, anomaly localization and detection are performed by comparing the reconstructed features $\bm{x}^{rec}_{0}$ with the original features $\bm{x}_{0}$. }
	\label{fig:framework}
\end{figure*}

Existing approaches for unsupervised anomaly detection and localization can be mainly divided into three categories: one-class classification based, feature embedding based, and reconstruction based.


One-class classification based methods build one-class classifiers from normal samples to distinguish them from anomalies.
One-class support vector machine (OC-SVM) \cite{SVM4AD,OC-SVM} and support vector data description (SVDD) \cite{SVDD,DeepSVDD,SemiSVDD} are the most widely employed models for anomaly detection.
Based on that, PatchSVDD \cite{PatchSVDD} further splits an input image into patches and adopts self-supervised learning for anomaly localization.

Feature embedding based approaches generally consist of two steps, feature extraction and anomaly estimation \cite{Tao2022DeepLF}. 
For feature extraction, recent work \cite{PANDA,PretrainedModeling,PretrainedRobustness} shows that using a network pre-trained on a large dataset like ImageNet \cite{ImageNet} is effective for AD.
For anomaly estimation, approaches like \cite{SPADE} compare deep feature embeddings of a target image and its K-nearest anomaly-free images.
PatchCore \cite{PatchCore} further introduces a memory bank for faster inference and reduced feature storage capacity.
Others
perform knowledge distillation to train a student model to estimate a scoring function \cite{MKD,RD4AD}.
Moreover, normal samples are also modelled with Gaussian distribution or self-organizing maps (SOM) and anomaly scores computed based on Mahalanobis distance \cite{PaDiM,SemiOrthog,SOMAD,GaussianFT}.

Reconstruction based methods are developed based on the hypothesis that a model trained to reconstruct normal samples only cannot correctly reconstruct anomalous samples \cite{VisualExplainingVAE,GMM4AD}. 
In this line of research, auto-encoder (AE) is one of the most widely used architecture for reconstruction \cite{fisrtANAE,AE4MRI,zhao2017spatio,DFR,VT-ADL}. 
Some work investigates the broader self-supervised learning paradigm for anomaly reconstruction \cite{DRAEM,CutPaste,InpaintingTransformer,RIAD,AttributeRestoration}.
UniAD \cite{UniAD} uses transformer as the backbone network and proposes a neighborhood mask encoder, a trainable query embedding, and a feature jittering strategy to address the shortcut issue.
Multiple works also constraint the representation of latent space via memory banks \cite{MemAE,niu2022memoryBank} and clustering \cite{yang2019clustering}. 
Another prevalent methodology is based on generative models.
Unlike auto-encoders that only consider the final reconstruction, generative model based studies aim to obtain the feature distribution of normal data.
Since generative models only learn how to generate normal samples and thus the anomalous region can be localized using the discrepancy between the generated or constructed sample and the input \cite{Tao2022DeepLF}.
Existing generative models for anomaly detection and localization primarily include variational auto-encoder (VAE) \cite{VisualExplainingVAE,AttenVAE,VAEsystem}, generative adversarial network (GAN) \cite{Ganomaly,fAnoGAN,CycleGANAD,ALOCC}.

In this paper, we build our framework based on generative models for their superior reconstruction performance of normal samples \cite{Tao2022DeepLF} and take the diffusion model as backbone.
The most related approaches to ours are \cite{AnoDDPMAD,FastUB}, which are also diffusion model based. 
However, both of them only target medical anomaly detection, while our approach considers more general cases ranging from texture level to semantic level.
Moreover, \cite{AnoDDPMAD} learns diffusion models at the image level and develops a multi-scale simplex noise diffusion process to control the target anomaly size which is particularly critical in the medical domain.
In contrast, our approach operates a latent diffusion model in the feature space for better scalability.
Compared with \cite{FastUB}, we additionally propose a memory bank based feature editing strategy to mitigate the shortcut issue of the latent diffusion model \cite{LDM}, and meanwhile, improve the reconstruction performance by pulling the initial state closer to the normal data manifold.
Compared with other reconstruction based approaches with memory bank \cite{MemAE,tan2021trustmae}, we produce the memory bank without training.
Besides, we operate the memory bank at the input stage of the reconstruction model rather than the compressed latent vector space.

\section{Method}
In this section, we elaborate on the proposed LafitE.
Figure \ref{fig:framework} illustrates the overall framework.




\subsection{Hierarchical Feature Extraction}
\label{sec:feature_extraction}

Let $\mathcal{X}_N = \{\bm{X}_{N}^{(i)}\}_{i=1}^M$ denote the set of normal samples for training, where $\bm{X}_N^{(i)}\in\mathbb{R}^{H\times W \times C}$ is the $i$-th image.
$H$, $W$, and $C$ denote height, width, and channels, respectively.
We use a pre-trained network
as the image feature extractor and adopt a multi-scale aggregation (MSA) module to integrate hierarchical features of the input image.

Let $\phi$ denote a convolutional neural network with $L$ levels.
We first encode an image $\bm{X} \in \mathbb{R}^{H\times W\times C}$ into a set of feature representations $\phi(\bm{X}) = \{ \phi_{1}(\bm{X}),\phi_{2}(\bm{X}),...,\phi_{L}(\bm{X}) \}$, where $\phi_{l}(\bm{X}) \in \mathbb{R}^{h_{l}\times w_{l}\times c_{l}}, l\in [1,L]$.
Considering that features from different layers correspond to different levels of information, \eg, shallow layers contain more texture information and deep layers express semantic information,
Here we select a subset of features $\mathcal{S}(\phi(\bm{X})) \subset \phi(\bm{X})$ and exclude the last layer to avoid the bias to the classification task for pre-training.


Then, we operate the MSA module: we perform feature alignment via an aggregation operation over $\mathcal{S}(\phi(\bm{X}))$.
We scale features from different layers in $\mathcal{S}(\phi(\bm{X}))$ to a unified spatial dimension of height as $h$ and width as $w$, and concatenate them along the channel axis with dimension $c$.
Denote the resulted feature tensor as $\bm{x}$, we have:
\begin{equation}
  \bm{x} = \text{Concat}(\{\text{Scale}(\phi_l(\bm{X}))\}),\quad\phi_l(\bm{X})\in \mathcal{S}(\phi(\bm{X})).
  \label{eq:1}
\end{equation}

Let $\bm{x}_{(i,j)}\in\mathbb{R}^c$ denote the feature slice at the position $(i,j)$ of the aggregated feature tensor $\bm{x}$, it represents the dense multi-scale representations of corresponding patch-level regions.
As the perceptual field of a patch embedding is usually large, the information contained in $\bm{x}_{(i,j)}$ may range from texture level to semantic level, thus benefiting the anomaly detection with various anomalous types in the multi-class setting.


\subsection{Diffusion Model for Normal Feature Modeling}

We utilize the denoising diffusion implicit model (DDIM) \cite{DDIM} in the latent space to model the compact distribution of normal samples.
Given a feature tensor of a normal example, we aim to gradually corrupt it into Gaussian white noise by the noise addition process, \ie, the diffusion process. 
Then, the network is trained to learn the score function of the normal distribution from Gaussian noise to normal feature tensors, which is called the denoising process, \textit{a.k.a.}, the reverse process.

Following \cite{DDPM}, we define the forward Markov diffusion process to generate images $\bm{x}_t$ with noises of different scale from the input $\bm{x}_0$:
\begin{equation}
  q(\bm{x}_{t}|\bm{x}_{t-1})=\mathcal{N}(\bm{x}_{t};\sqrt{1-\beta_{t}}\bm{x}_{t-1},\beta_{t}\bm{I}),
  \label{eq:2}
\end{equation}
where $0<\beta_{t}<1$ is the variance of noise level $t$, which can be also regarded as moment $t$.
Larger $t$ implies more noise.
The marginal distribution of this Markov chain at moment $t$ can be explicitly written as:
\begin{equation}
  q(\bm{x}_{t}|\bm{x}_{0})=\mathcal{N}(\bm{x}_{t};\sqrt{\bar{\alpha}_{t}}\bm{x}_{0},\sqrt{1-\bar{{\alpha}}_{t}}\bm{I}),
  \label{eq:3}
\end{equation}
with $\alpha_{t} = 1-\beta_{t}$ and $\bar{\alpha}_{t}=\prod^{t}_{s=1}(1-\beta_{s})$.

To generate samples from Gaussian noise $\bm{x}_T$ in the reverse processes, DDIM models the stepwise denoising prior probability as a Gaussian distribution as well:
\begin{equation}
  p_{\theta}(\bm{x}_{t-1}|\bm{x}_{t})=\mathcal{N}(\bm{x}_{t-1};\mu_{\theta}(\bm{x}_{t},t), \sigma^{2}_{t}\bm{I}),
  \label{eq:4}
\end{equation}
where $\sigma_{t}$ is fixed for each $t$ and the mean functions $\mu_{\theta}(\bm{x}_{t},t)$ are trainable, which can be formulated as:
\begin{equation}
  \mu_{\theta}(\bm{x}_{t},t) = \frac{1}{\alpha_{t}}(\bm{x}_{t}-\frac{\beta_{t}}{\sqrt{1-\bar{\alpha}_{t}}} \bm{\epsilon}_{\theta}(\bm{x}_{t},t)).
  \label{eq:5}
\end{equation}
$\bm{\epsilon}_{\theta}$ is the noise function, (or score function), which can be predicted with the network parameterized by $\theta$.
Here we adopt U-net \cite{U-net} as the network architecture, since its skip connections can effectively avoid irreversible information reduction brought by down-sampling in other CNN based auto-encoders.
The training objective can be written as:
\begin{equation}
  \mathcal{L}_{\theta} = \mathbb{E}_{\bm{x},\bm{\epsilon}}|| \bm{\epsilon} - \bm{\epsilon}_{\theta}(\sqrt{\bar{{\alpha}_{t}}}\bm{x}_{0} + \sqrt{1-\bar{{\alpha}}_{t}}\bm{\epsilon} , t)||^2_{2},
  \label{eq:6}
\end{equation}
where $\bm{\epsilon} \sim \mathcal{N}(0,\bm{I})$.

Following DDIM \cite{DDIM}, we perform iterative sampling in the denoising generation process via the following equation:
\begin{equation}
\begin{aligned}
  \bm{x}_{t-\Delta t} &= \sqrt{\bar{\alpha}_{t-\Delta t}}(\frac{\bm{x}_{t}-\sqrt{1-\bar{\alpha}_{t}}\bm{\epsilon}_{\theta}(\bm{x}_{t},t)}{\bar{\alpha}_{t}}) \\
  &+ \sqrt{1-\bar{\alpha}_{t-\Delta t}-\sigma^{2}_{t}}\bm{\epsilon}_{\theta}(\bm{x}_{t},t) + \sigma_{t}\bm{z}_{t},
  \label{eq:7}
\end{aligned}
\end{equation}
with $\sigma_{t}=\eta \sqrt{(1-\bar{\alpha}_{t-\Delta t})/(1-\bar{\alpha}_{t})} \sqrt{1-\bar{\alpha}_{t}/\bar{\alpha}_{t-\Delta t}}$, $\bm{z}_{t} \sim \mathcal{N}(0,\bm{I})$, and $\Delta t$ denoting the sampling interval.
With the trained diffusion model, we can generate the feature tensor of a normal sample from its corrupted noisy features by running the denoising process.

\subsection{Reconstruction with Feature Editing}
\paragraph{Diffusion Reconstruction} We leverage the diffusion model as the backbone of the reconstruction framework.
Instead of corrupting the original features into the Gaussian noise during the diffusion process,
we pre-define a maximum corruption step $\tau \le T$ to obtain the initial states for the reverse process during inference.
In this way, the resulted $\bm{x}_{\tau}$ still keeps some information about the input $\bm{x}_0$, and we can recursively sample from this partially noised sample for reconstruction.
The corrupted feature tensor can be denoted as:
\begin{equation}
    \bm{x}_{\tau}=\sqrt{\bar{\alpha}_{\tau}}\bm{x}_{0}+\sqrt{1-\bar{{\alpha}}_{\tau}}\bm{\epsilon}.
  \label{eq:9}
\end{equation}

Equation (\ref{eq:9}) shows that the initial state for reconstruction, \ie, $\bm{x}_{\tau}$ inevitably contains the information in $\bm{x}_{0}$.
However, more information from $\bm{x}_{0}$ also means a higher possibility of identity-shortcut \textit{w.r.t.} the anomalous regions. 
Therefore, we hope to find an appropriate $\tau$, so that the corrupted $\bm{x}_{\tau}$ contains necessary information to reconstruct the normal regions in $\bm{x}_0$, while the anomalous information is dispersed by the added noise.

\paragraph{Feature Editing}
We first build up a memory bank to preserve the representations of normal samples:
\begin{equation}
  \mathcal{M}=\bigcup\limits_{\substack{\bm{x} \in \mathcal{X}_{N}, i\in h, j
  \in w}} \bm{x}_{(i,j)}.
  \label{eq:8}
\end{equation}

Since the size of $\mathcal{M}$ is proportional to that of the training data, we sample a core set $\mathcal{M}_C$ from the whole memory bank $\mathcal{M}$ with a greedy search algorithm\cite{agarwal2005geometric} to diminish information redundancy, reduce storage space, and significantly improve inference speed.

Then, we edit each feature slice $\bm{x}^Q_{(i,j)}$ of a query sample $\bm{x}^Q$ with a linear combination of its top-K nearest neighbors $\mathcal{N}_K(\bm{x}^Q_{(i,j)})=\{\bm{n}_k\}_{k=1}^K$ in the core set $\mathcal{M}_{C}$:
\begin{equation}
    \bm{e}_{i,j} = \sum\limits^{K}_{k=1} \left(1 - \frac{\exp||\bm{x}^Q_{(i,j)}-\bm{n}_{k}||_{2}}{\sum_{\bm{n}\in \mathcal{N}_K(\bm{x}^Q_{(i,j)})} \exp||\bm{x}^Q_{(i,j)}-\bm{n}||_{2}} \right) \cdot \bm{n}_{k},
  \label{eq:10}
\end{equation}
where $\bm{e}_{(i,j)}$ the edited feature slice \textit{w.r.t.} at the position $(i,j)$ and the whole feature tensor after editing can be denoted as $\bm{e}$.
The weighting term for $\bm{n}_k$ facilitate the edited feature tensor to integrate the information of similar, but different, normal patterns stored in $\mathcal{M}_C$, and meanwhile, taking into account the overall characteristics of normal data.
More importantly, it also ensures that the distribution of edited features will not lie too far from the normal data manifold, which is conducive to the robustness of the method.




Note that we conduct the diffusion reconstruction on edited feature tensors $\bm{e}$,
which excludes abnormal information in advance to address the identity shortcut issue.
More specifically, the initial state for reconstruction is derived from:
\begin{equation}
  \bm{x}_{\tau}=\sqrt{\bar{\alpha}_{\tau}}\bm{e}+\sqrt{1-\bar{{\alpha}}_{\tau}}\bm{\epsilon}.
  \label{eq:11}
\end{equation}

So far, the editing is independently conducted on each feature slice and do not consider the spatial correlation among different feature slices.
Consequently, the joint distribution $p(\bm{e})$ has a certain distance from the normal feature distribution $p(\bm{x})$. 
Fortunately, applying a sufficient large noise on the edited feature in the diffusion process can make the manifold of $\bm{e}_{\tau}$ closer to that of $\bm{x}_{\tau}$, which enables us to reconstruct higher quality features via the reverse or denoising process.
Moreover, such reconstruction process considers the spatial correlation of feature slices in $\bm{e}$ via the joint distribution $p(\bm{e}_{\tau})$, which also contributes to better reconstruction performance. 


\paragraph{Anomaly Estimation}
With the reconstructed features from the denoising process, 
we finally drive the feature-level anomaly score map by computing L2-norm reconstruction error between the denoised feature tensor $\bm{x}_0^{rec}$ and the original $\bm{x}_0$.
Then we up-sample the score map with bi-linear interpolation and apply Gaussian kernel smoothing to obtain the image-level score map for anomaly localization:
\begin{equation}
    \bm{s} = \text{Upsample}\left(||\bm{x}^{rec}_{0}-\bm{x}_{0}||^{2}_{2}\right) \ast g_{\sigma},
  \label{eq:12}
\end{equation}
where $\bm{s} \in \mathbb{R}^{H\times W}$, $g_{\sigma}$ denotes the Gaussian kernel with standard deviation of $\sigma$, and $*$ denotes the convolution operator.
We perform average pooling on $\bm{s}$ and take the maximum value as the final image-level anomaly score.

\begin{figure}[t]
    \centering
     \hspace{0.005\linewidth}
     \begin{subfigure}[]{0.2\linewidth}
         \centering
         \includegraphics[width=\linewidth]{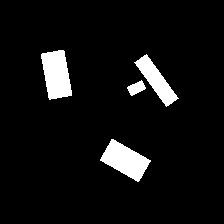}
     \end{subfigure}
     \begin{subfigure}[]{0.2\linewidth}
         \centering
         \includegraphics[width=\linewidth]{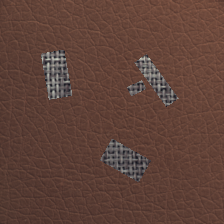}
     \end{subfigure}
     \hspace{0.01\linewidth}
     \begin{subfigure}[]{0.2\linewidth}
         \centering
         \includegraphics[width=\linewidth]{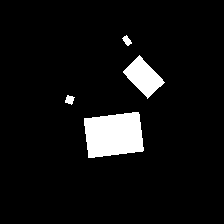}
     \end{subfigure}
     \begin{subfigure}[]{0.2\linewidth}
         \centering
         \includegraphics[width=\linewidth]{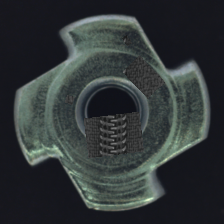}
     \end{subfigure}
     
    \caption{
        Masks and the synthesized anomalous samples.
    }
    \label{fig:pseudo_eg}
\end{figure}

\subsection{Anomalous Data Synthesis}
\label{pseudo Anomaly Data Synthesizing}
Since normal samples are the only ones accessible in unsupervised anomaly detection, we found an absence of a validation set to guide the hyperparameter selection scheme.
Such an issue caused by the intrinsic nature of the task has not been explored at either the concept level or experiment level in previous work.
Inspired by CutPaste \cite{CutPaste}, we propose a straightforward method to synthesize anomaly samples and then build a pseudo validation set for hyperparameter selection. 

Specifically, given a set of normal samples, we first initiate multiple masks of different shapes and sizes to ensure diversity.
Then, we randomly rotate and combine these masks to produce a set of fragments from each normal sample.
After that, we synthesize the anomaly sample by sticking these extracted fragments to another randomly selected normal sample, and these fragments are labeled as anomalous regions.
Figure \ref{fig:pseudo_eg} is the exemplification.


\begin{table*}[!t]
    \centering
    \setlength{\tabcolsep}{1mm}
    \caption{AUROC scores of multi-class anomaly \textbf{detection} and \textbf{localization} (shown as Det. / Loc.) on MVTec-AD \cite{MVTec}. We highlight the highest results in color and underline the results lower than 80.0\%.
    }
    \resizebox{0.99\textwidth}{!}{
    \begin{tabular}{c|ccccccc|ccc}
    \toprule
         \textbf{Category} 
         & \textbf{PatchSVDD} \cite{PatchSVDD} 
         & \textbf{PaDiM} \cite{PaDiM} 
         & \textbf{DRAEM} \cite{DRAEM} 
         & \textbf{RevDistill} \cite{RD4AD} 
         & \textbf{PatchCore} \cite{PatchCore} 
         & \textbf{FastFlow} \cite{yu2021fastflow} 
         & \textbf{UniAD} \cite{UniAD} 
         & \multicolumn{3}{c}{\textbf{LafitE (ours)}} \\
         \midrule
          Bottle 
          & 85.5 / 87.6
          & 97.9 / 96.1
          & 97.5 / 87.6
          & \underline{46.2} / 95.5  
          & 97.5 / 98.1
          & 97.6 / 84.9
          & 99.7 / 98.1
          & \textbf{\textcolor{red}{100}} \small{$\pm$ 0.00} \normalsize 
                 & / & \textbf{\textcolor{green}{98.4}} \small{$\pm$ 0.01}
          \\
           Cable 
           & \underline{64.4} /  \underline{62.2}
           & \underline{70.9} /  81.0
           & \underline{57.8} /  \underline{71.3}
           & \underline{76.9} / 83.4
           & \textbf{\textcolor{red}{99.7}} /  \textbf{\textcolor{green}{98.1}}
           & 97.9 /  97.4
           & 95.2 /  97.3
           & 97.2 \small{$\pm$ 0.25} \normalsize 
                    & / & 97.3 \small{$\pm$ 0.07}
          \\
           Capsule 
           & \underline{61.3} / 83.1
           & \underline{73.4} /  96.9
           & \underline{65.3} /  \underline{50.5}
           & \textbf{\textcolor{red}{97.3}} / 98.5
           & \underline{78.0} / 97.6
           & 95.6 /  98.7
           & 86.9 / 98.5
           & 96.8 \small{$\pm$ 0.25} \normalsize 
                    & / & \textbf{\textcolor{green}{98.9}} \small{$\pm$ 0.01}
          \\
           Hazelnut 
           & 83.9 / 97.4
           & 85.5 / 96.3
           & 93.7 / 96.9
           & \textbf{\textcolor{red}{100}} / 98.9
           & \textbf{\textcolor{red}{100}} / 98.4
           & 98.2 / \textbf{\textcolor{green}{99.0}}
           & 99.8 / 98.1
           & \textbf{\textcolor{red}{100}} \small{$\pm$ 0.00} \normalsize 
                    & / & 98.4 \small{$\pm$ 0.05}
          \\
           Metal Nut 
           & 80.9 / 96.0
           & 88.0 /  84.8
           & \underline{72.8} / \underline{62.2}
           & 94.9 /  93.4
           & 98.6 / \textbf{\textcolor{green}{98.5}}
           & 98.2 / 97.7
           & 99.2 / 94.8
           & \textbf{\textcolor{red}{99.8}} \small{$\pm$ 0.06} \normalsize 
                    & / & 96.8 \small{$\pm$ 0.01}
          \\
           Pill  
           & 89.4 /  96.5
           & \underline{68.8} / 87.7
           & 82.2 / 94.4
           & 94.9 / 98.3
           & 94.8 / 97.6
           & 95.0 / \textbf{\textcolor{green}{98.7}}
           & 93.7 / 95.0
           & \textbf{\textcolor{red}{98.0}} \small{$\pm$ 0.16} \normalsize 
                    & / & 96.7 \small{$\pm$ 0.03}
          \\
           Screw 
           & 80.9 /  \underline{74.3}
           & \underline{56.9} / 94.1
           & 92.0 / 95.5
           & \textbf{\textcolor{red}{96.5}} / 99.4
           & \underline{72.0} / 96.5
           & \underline{79.6} / 98.3
           & 87.5 / 98.3
           & 95.4 \small{$\pm$ 0.28} \normalsize 
                    & / & \textbf{\textcolor{green}{99.5}} \small{$\pm$ 0.01}
          \\
          Toothbrush 
          & \textbf{\textcolor{red}{99.4}} / 98.0
          & 95.3 / 95.6
          & 90.6 / 97.7
          & 83.6 /  \textbf{\textcolor{green}{98.9}}
          & 98.3 / 98.6
          & 81.9 / 98.7
          & 94.2 / 98.4
          & 92.9 \small{$\pm$ 0.22} \normalsize 
                    & / & 98.8 \small{$\pm$ 0.00}
          \\
          Transistor 
          & \underline{77.5} / \underline{78.5}
          & 86.6 / 92.3
          & \underline{74.8} / \underline{64.5}
          & 91.5 / 87.2
          & 99.6 / 97.3
          & 91.4 / 94.7
          & \textbf{\textcolor{red}{99.8}} / \textbf{\textcolor{green}{97.9}}
          & 99.6 \small{$\pm$ 0.10} \normalsize 
                    & / & 97.4 \small{$\pm$ 0.02}
          \\
          Zipper 
          & \underline{77.8} / 95.1
          & \underline{79.7} / 94.8
          & 98.8 / \textbf{\textcolor{green}{98.3}}
          & 98.9 / 98.1
          & 95.9 / 97.5
          & 95.4 / 97.9
          & 95.8 / 96.8
          & \textbf{\textcolor{red}{99.4}} \small{$\pm$ 0.09} \normalsize 
                    & / & \textbf{\textcolor{green}{98.3}} \small{$\pm$ 0.02}
          \\
          Carpet 
          & \underline{63.3} / \underline{78.6}
          & 93.8 / 97.6
          & 98.0 /  98.6
          & 95.7 / \textbf{\textcolor{green}{98.7}}
          & 96.8 / \textbf{\textcolor{green}{98.7}}
          & 99.6 / 94.0
          & \textbf{\textcolor{red}{99.8}} /  98.5
          & \textbf{\textcolor{red}{99.8}} \small{$\pm$ 0.05} \normalsize 
                    & / & \textbf{\textcolor{green}{98.7}} \small{$\pm$ 0.03}
          \\
          Grid 
          & \underline{66.0} / \underline{70.8}
          & \underline{73.9} / \underline{71.0}
          & 99.3 / 98.7
          & 97.9 / \textbf{\textcolor{green}{99.2}}
          & 82.6 / 96.6
          & 91.1 / 96.3
          & 98.2 / 96.5
          & \textbf{\textcolor{red}{99.9}} \small{$\pm$ 0.03} \normalsize 
                    & / & 98.5 \small{$\pm$ 0.01}
          \\
          Leather 
          & \underline{60.8} / 93.5
          & 99.9 / 84.8
          & 98.7 / 97.3
          & \textbf{\textcolor{red}{100}} / \textbf{\textcolor{green}{99.3}}
          & \textbf{\textcolor{red}{100}} / 99.0
          & 95.8 / 92.6
          & \textbf{\textcolor{red}{100}} / 98.8
          & \textbf{\textcolor{red}{100}} \small{$\pm$ 0.00} \normalsize 
                    & / & 99.2 \small{$\pm$ 0.02}
          \\
          Tile
          & 88.3 / 92.1
          & 93.3 / 80.5
          & 99.8 / \textbf{\textcolor{green}{98.0}}
          & 97.7 / 95.8
          & 98.4 / 94.8
          & 99.8 / 93.5
          & 99.3 / 91.8
          & \textbf{\textcolor{red}{100}} \small{$\pm$ 0.00} \normalsize 
                    & / & 93.2 \small{$\pm$ 0.04}
          \\
          Wood 
          & \underline{72.1} / 80.7
          & 98.4 / 89.1
          & \textbf{\textcolor{red}{99.8}} / \textbf{\textcolor{green}{96.0}}
          & 98.9 / 95.8
          & 97.1 / 93.8
          & 98.3 / 91.8
          & 98.6 / 93.2
          & 98.7 \small{$\pm$ 0.18} \normalsize 
                    & / & 94.3 \small{$\pm$ 0.15}
          \\
          \midrule
          \rowcolor{gray!10}
          \textbf{Average}
          & \underline{76.8} / 85.6
          & 84.2 / 89.5
          & 88.1 / 87.2
          & 91.4 / 96.0
          & 94.0 / 97.4
          & 94.4 / 95.6
          & 96.5 / 96.8
          & \textbf{\textcolor{red}{98.5}} \small{$\pm$ 0.03} \normalsize 
                     & / & \textbf{\textcolor{green}{97.6}} \small{$\pm$ 0.02}
          \\
          \midrule
          \rowcolor{gray!10}
          \textbf{Inter-category Std}
          & 11.35 / 11.14
          & 12.54 / 7.56
          & 13.54 / 16.36
          & 13.57 / 4.75
          & 8.56  / \textbf{\textcolor{green}{1.42}}
          & 5.90  / 3.72
          & 4.21  / 2.12
          & \textbf{\textcolor{red}{2.04} }
                     & / & 1.73
          \\
        \bottomrule     
    \end{tabular}
    }
    \label{tab:AD-MVtec}
\end{table*}

\section{Experiment}
\subsection{Experimental Setup}
\paragraph{Dataset.}
We conduct experiments on two widely used datasets, MVTec-AD \cite{MVTec} and MPDD \cite{MPDD}, to validate our approach.
Both datasets have image-level labels and pixel-level annotations.
MVTec-AD consists of 3629 images for training and validation, and 1725 images for testing.
The training set contains only normal images, while the test dataset contains both normal and anomalous images.
There are 15 real-world categories in this dataset, including 5 classes of textures and 10 classes of objects, and each class has multiple types of defects.
MPDD contains 6 classes of metal parts, focusing on defect detection during the fabrication of painted metal parts.
Its training set is composed of 888 normal samples without defects, and the test set is composed of 458 samples either normal or anomalous.
In particular, samples in MPDD have non-homogeneous backgrounds with diverse spatial orientations, different positions, and various light intensities, 
leading to greater challenges in anomaly detection.

\begin{table*}[!t]
    \centering
    \setlength{\tabcolsep}{1mm}
    \caption{
    AUROC scores of multi-class anomaly \textbf{detection} and \textbf{localization} (shown as Det. / Loc.) on MPDD \cite{MPDD}. We highlight the highest results in color and underline the results lower than 80.0\%.
    }
    \resizebox{0.99\textwidth}{!}{
    \begin{tabular}{c|ccccccc|ccc}
    \toprule
         \textbf{Category} 
         & \textbf{PatchSVDD} \cite{PatchSVDD} 
         & \textbf{PaDiM} \cite{PaDiM} 
         & \textbf{DRAEM} \cite{DRAEM} 
         & \textbf{RevDistill} \cite{RD4AD} 
         & \textbf{PatchCore} \cite{PatchCore} 
         & \textbf{FastFlow} \cite{yu2021fastflow} 
         & \textbf{UniAD} \cite{UniAD} 
         & \multicolumn{3}{c}{\textbf{LafitE (ours)}} \\
         \midrule
          Bracket Black  
          & 85.8 / \underline{67.9}
          & \underline{71.1} / 93.1
          & 81.2 / 97.9
          & 81.0 / 97.3
          & \underline{77.3} / 96.9
          & 81.4 / 82.4
          & 96.8 / 94.9
          & \textbf{\textcolor{red}{98.5}} \small{$\pm$ 0.12} \normalsize 
                     & / & \textbf{\textcolor{green}{99.3}} \small{$\pm$ 0.02}
          \\
          Bracket Brown  
          & 97.3 / \underline{63.2}
          & \underline{75.0} / 95.0
          & 85.0 / \underline{53.8}
          & 86.0 / 97.2 
          & 83.1 / 95.3
          & 97.5 / 80.3
          & \textbf{\textcolor{red}{98.9}} / 98.6
          & 96.5 \small{$\pm$ 0.39} \normalsize 
                     & / & \textbf{\textcolor{green}{99.5}} \small{$\pm$ 0.01}
          \\
          Bracket White  
          & 87.2 / \underline{55.8}
          & \underline{73.0} / 97.2
          & \underline{78.8} / 95.7
          & 83.6 / 98.8
          & \underline{75.8} / \textbf{\textcolor{green}{99.6}}
          & \underline{72.3} / 98.1
          & 88.7 / 96.0
          & \textbf{\textcolor{red}{92.4}} \small{$\pm$ 0.39} \normalsize 
                     & / & 98.4 \small{$\pm$ 0.01}
          \\
          Connector  
          & \textbf{\textcolor{red}{99.8}} / 90.2
          & 83.8 / 97.2
          & 88.8 / 85.1
          & 99.5 / \textbf{\textcolor{green}{99.5}}
          & 96.4 / 98.4
          & 94.0 / 94.0
          & 91.0 / 98.0
          & 96.7 \small{$\pm$ 0.23} \normalsize 
                     & / & 99.1 \small{$\pm$ 0.02}
          \\
          Metal Plate  
          & 84.6 / 91.0
          & \underline{51.1} / 90.2
          & \textbf{\textcolor{red}{100}} / \textbf{\textcolor{green}{99.2}}
          & \textbf{\textcolor{red}{100}} / \textbf{\textcolor{green}{99.2}}
          & \textbf{\textcolor{red}{100}} / 98.6
          & 99.7 / 97.9
          & \underline{73.0} / 94.0
          & \textbf{\textcolor{red}{100}} \small{$\pm$ 0.00} \normalsize 
                     & / & 98.7 \small{$\pm$ 0.01}
          \\
          Tubes  
          & \underline{79.1} / \underline{41.7}
          & \underline{75.6} / 88.7
          & \textbf{\textcolor{red}{96.2}} / 98.2
          & 95.5 / 99.1
          & \underline{68.5} / 97.3
          & \underline{77.1} / 96.9
          & \underline{76.9} / 92.3
          & 94.8 \small{$\pm$ 0.46} \normalsize 
                     & / & \textbf{\textcolor{green}{99.2}} \small{$\pm$ 0.01}
          \\

          \midrule
          \rowcolor{gray!10}
          \textbf{Average}
          & 89.0 / \underline{68.3}
          & \underline{71.6} / 93.6
          & 88.3 / 88.3
          & 90.9 / 98.5
          & 83.5 / 97.7
          & 87.0 / 91.6
          & 87.5 / 95.6
          & \textbf{\textcolor{red}{96.5}} \small{$\pm$ 0.08} \normalsize 
                     & / & \textbf{\textcolor{green}{99.0}} \small{$\pm$ 0.01}
          \\
          \midrule
          \rowcolor{gray!10}
          \textbf{Inter-category Std} 
          & 7.26 / 17.72
          & 9.99 / 3.26
          & 7.66 / 16.14
          & 7.67 / 0.92
          & 11.28 / 1.38
          & 10.52 / 7.40
          & 9.59 / 2.19
          & \textbf{\textcolor{red}{2.44}}
                     & / & \textbf{\textcolor{green}{0.36}}
          \\
        \bottomrule     
    \end{tabular}
    }
    \label{tab:AD-MPDD}
\end{table*}

\paragraph{Implementation Details.}
For experiments on both MVTec-AD and MPDD, we use bi-linear interpolation to resize the images to $224 \times 224$ without any other enhancement. 
We adopt EfficientNet-b4 \cite{tan2019efficientnet} pre-trained on ImageNet as $\phi$ in Section \ref{sec:feature_extraction}.
Feature maps from stage 1 to stage 4 are resized to $32 \times 32$ for feature aggregation following UniAD \cite{UniAD}.
We apply U-net \cite{U-net} for $\epsilon_\theta$ in the diffusion model.
The base channel is set to 256 and channels multiple is set to (1, 2, 3, 4).
We train the diffusion model for 1000 epochs on a single GPU (NVIDIA GeForce RTX 3090 24GB) with a batch size of 64, maximum time step $T$ of 1000, and the cosine noise schedule \cite{nichol2021improved}.
Following UniAD \cite{UniAD}, we use the AdamW optimizer \cite{loshchilov2017decoupled} with weight decay $1 \times 10^{-4}$, the learning rate is set to $1 \times 10^{-4}$ initially and dropped by 0.1 after 800 epochs.
We use a keep rate of 10\% to build the core set of the memory bank. 
Deviation for $g_\theta$ in Equation (\ref{eq:12}) is set to 4.
For the setting of the corruption step $\tau$ and the number of neighbors $K$ in Equation (\ref{eq:10}), please refer to Section \ref{sec:eval_hyperparam_selection}.
All results are reported from 5 random seeds.

\subsection{Performance Comparison}
In this section, we compare the proposed LafitE with various baseline models including one-class classification based PatchSVDD \cite{PatchSVDD}; feature-embedding based PaDiM \cite{PaDiM}, RevDistill \cite{RD4AD}, DRAEM \cite{DRAEM}, and PatchCore \cite{PatchCore}; auto-encoder based UniAD \cite{UniAD}; and generative model based FastFlow \cite{yu2021fastflow}.
Numbers for PatchSVDD \cite{PatchSVDD}, PaDiM \cite{PaDiM}, DRAEM \cite{DRAEM}, and UniAD \cite{UniAD} are copied from UniAD \cite{UniAD} for fair comparison.
As for other results of MVTec-AD and MPDD \cite{MPDD}, 
we reproduced them in this work with publicly available implementations.
Tables \ref{tab:AD-MVtec}, \ref{tab:AD-MPDD}, and \ref{tab:aupr} report the results of difference approaches on MVTec-AD and MPDD for multi-class anomaly detection and localization, respectively.
We can draw observations as follows.

\textbf{The proposed LafitE surpasses previous state-of-the-art methods in both performance and robustness across various metrics and datasets with different levels of complexity.}
On MVTec-AD, compared with UniAD, the previous state-of-the-art, 
our method obtains significant improvements of detection AUROC from 96.5\% to 98.5\% on average. 
LafitE also shows superiority over UniAD for localization AUROC with an average of 0.8\%.
On MPDD, our method outperforms RevDistill by 5.6\% and 0.5\% in terms of detection AUROC and localization AUROC, respectively.
This well demonstrates the effectiveness of the proposed approach.
Considering that the amount of normal and abnormal samples are often unbalanced in real application scenarios, the metric AU-PR was exploited to have a fairer view on performance evaluation \cite{zou2022spot}.
Thus we also evaluate AU-PR on both datasets here and Table \ref{tab:aupr} shows the results.
Generally speaking, LafitE achieves superior performance.
Though the performance of PatchCore on MVTec-AD anomaly localization looks promising, it drops dramatically on MPDD, where the normal samples are more complicated.
By comparison, our approach demonstrates more robustness.

\textbf{Our method demonstrates less bias to the category of objects and leads to more consistent performance from anomaly detection to anomaly localization.}
As shown in Table \ref{tab:AD-MVtec} and \ref{tab:AD-MPDD}, though prior methods such as RevDistill \cite{RD4AD}, PatchCore \cite{PatchCore}, FastFlow \cite{yu2021fastflow}, and UniAD \cite{UniAD} achieve promising averaged performance across all classes, their inter-category standard variation is much larger than ours with a maximum gap of 11.53, highlighting the consistency of our approach for multi-class anomaly detection.
Table \ref{tab:AD-MVtec} and \ref{tab:AD-MPDD} also indicate that unlike RevDistill \cite{RD4AD} and PatchCore \cite{PatchCore} producing excellent performance for localization but resulting in unsatisfied results for detection, our model achieves the best averaged performance for both detection and localization.

\textbf{Reconstruction based approaches such as LafitE and UniAD \cite{UniAD} generally lead to superior detection performance than feature embedding based approaches like RevDistill \cite{RD4AD} and PatchCore \cite{PatchCore}.}
We conjecture this is because feature editing based methods focus more on local patches while the nature of reconstruction makes the reconstruction based approaches gain a global view of inputs.
This awareness of the relationship among adjacent patches enhances the ability to deal with image level detection.

\begin{table}[!t]
    \centering
    \caption{Averaged AU-PR on MVTec-AD \cite{MVTec} and MPDD \cite{MPDD}. Best results are in bold.}
    \resizebox{0.47\textwidth}{!}{
    \begin{tabular}{l|cc|cc}
        \toprule
        \multirow{2}{*}{\textbf{Method}} & \multicolumn{2}{c|}{\textbf{MVTec-AD}} & \multicolumn{2}{c}{\textbf{MPDD}} \\
        \cmidrule{2-5}
        & Det. & Loc. & Det. & Loc. \\
        \midrule
        RevDistill   & 95.3 & 50.1 & 92.3 & 41.6  \\
        PatchCore    & 98.2 & \textbf{54.3} & 90.4 & 36.8  \\
        FastFlow     & 97.0 & 42.6 & 80.1 & 27.8  \\
        UniAD        & 98.9 & 44.8 & 89.8 & 20.3  \\
        \midrule
        LafitE (ours)   & \textbf{99.4}  \small{$\pm$ 0.02}
                        & 51.3  \small{$\pm$ 0.16}
                        & \textbf{94.8} \small{$\pm$ 0.03}
                        & \textbf{44.9} \small{$\pm$ 0.22}
                        \\
    \bottomrule
    \end{tabular}
    }
    \label{tab:aupr}
\end{table}

\subsection{Ablation Study}

Here we validate the contributions of different components in the proposed LafitE.
We introduce two variants:
i) \textit{LafitE w/o F.E.}, which removes the feature editing strategy from the proposed LafitE and directly leverages the latent diffusion model for anomaly detection.
ii) \textit{UniAD w/ F.E.}, where we replace the diffusion model in LafitE with UniAD for reconstruction.
Table \ref{tab:ablation} highlights the results.
\begin{table}[!t]
    \centering
    \caption{Ablation results of averaged AUROC on MVTec-AD and MPDD. LDM denotes the latent diffusion model based backbone. F.E. denotes the feature editing strategy.}
    \resizebox{0.47\textwidth}{!}{
    \begin{tabular}{l|l|cc|c|c}
    \toprule
          & \textbf{Method} & \textbf{Architecture} & \textbf{F.E.} & \textbf{Detection} & \textbf{Localization}\\ 
         \midrule
         \multirow{4}{*}{\rotatebox{90}{MVTec-AD}}
         & LafitE & LDM & \usym{2713} & \textbf{98.5} & \textbf{97.6} \\
         & LafitE w/o F.E. & LDM & \usym{2717} & 98.2 & 97.4 \\
         & UniAD w/ F.E. & Transformer & \usym{2713} & 96.8 & 97.1 \\
         & UniAD & Transformer & \usym{2717} & 96.5 & 96.8 \\
         \midrule
         \multirow{4}{*}{\rotatebox{90}{MPDD}}
         & LafitE & LDM & \usym{2713} & \textbf{96.5} & \textbf{99.0} \\
         & LafitE w/o F.E. & LDM & \usym{2717} & 93.6 & 98.3 \\
         & UniAD w/ F.E. & Transformer & \usym{2713} & 86.6 & 95.6 \\
         & UniAD & Transformer & \usym{2717} & 87.5 & 95.6 \\
    \bottomrule
    \end{tabular}
    }
    \label{tab:ablation}
\end{table}


Comparing LafitE with \textit{LafitE w/o F.E.}
,  we can see that incorporating the proposed feature editing strategy brings consistent performance gain to the basic construction model on both datasets, well verifying its effectiveness.
Table \ref{tab:ablation} also shows that LafitE outperforms \textit{UniAD w/ F.E.} and \textit{LafitE w/o F.E.} outperforms UniAD.
This highlights the superiority of the proposed generative model based approach over the auto-encoder framework in UniAD.

We further visualize the reconstructed images by UniAD, \textit{LafitE w/o F.E.}, and LafitE for more in-depth understanding.
As shown in Figure \ref{fig:motivation}, though UniAD well reconstructs normal regions, anomalous information is still kept in the reconstructed image, thus hurting the distinguishability between normal regions and anomalous regions for anomaly detection.
In contrast, our LafitE and \textit{LafitE w/o F.E.} exhibit much less identity shortcut issues in anomalous regions.
Figure \ref{fig:motivation} also depicts that compared with \textit{LafitE w/o F.E.}, LafitE not only produces larger discrepancy in anomalous regions after reconstruction which alleviates the identity shortcut issue, but also leads to lower reconstruction error in normal regions which helps reduce the false positive predictions.
This implies that the proposed feature editing strategy can benefit the diffusion model with the corrupted states getting closer to the normal data manifold, well verifying our motivation.

\subsection{Feature Distribution in Diffusion Process}
\begin{figure}[hp]
	\centering
        \begin{subfigure}[]{0.3\linewidth}
             \centering
             \includegraphics[width=0.95\linewidth]{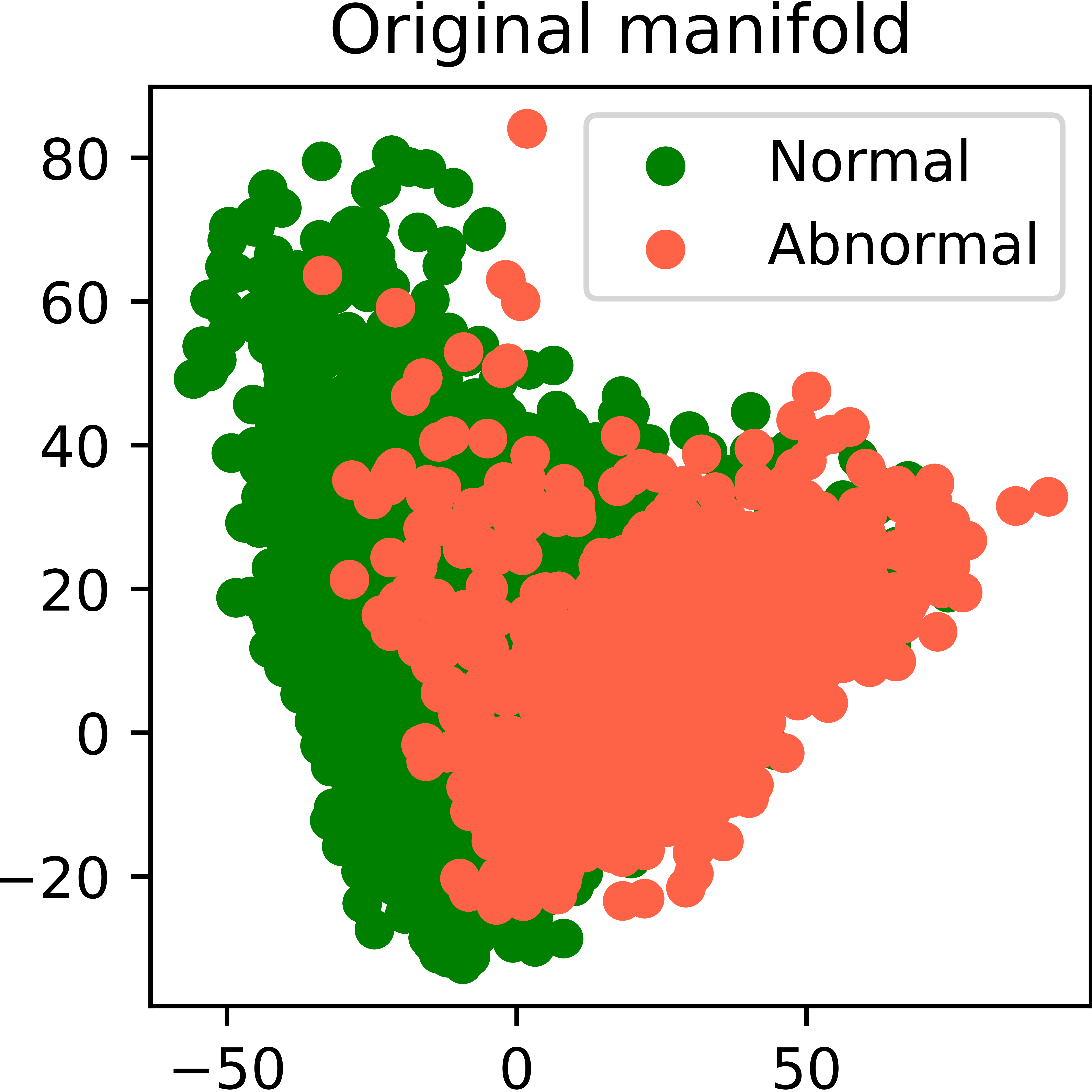}
             \caption{$\tau$=0}  \label{q1_0}
         \end{subfigure} 
         \hfill
        \begin{subfigure}[]{0.3\linewidth}
             \centering
             \includegraphics[width=0.95\linewidth]{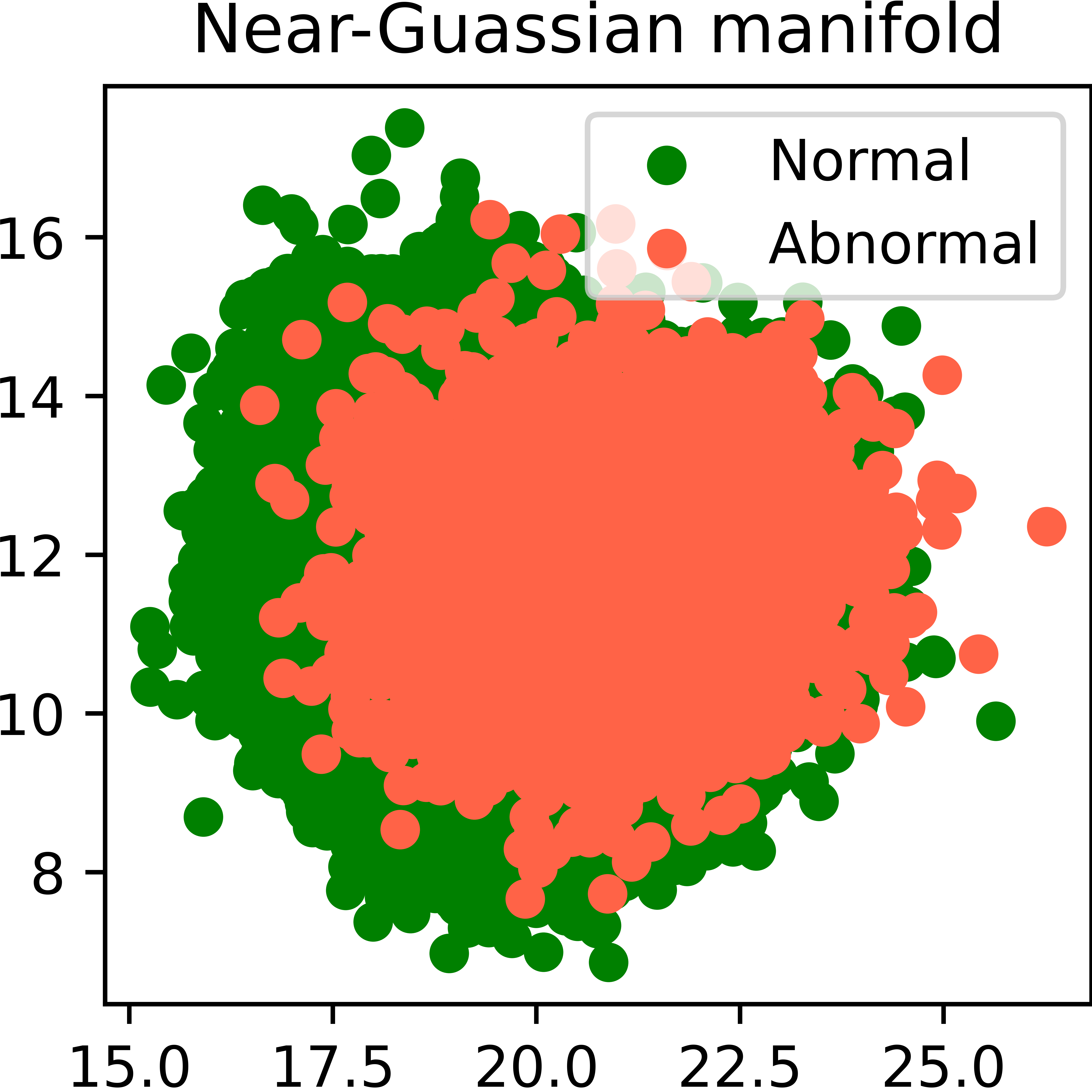}            
             \caption{$\tau$=970}  \label{q1_1}
         \end{subfigure}
         \hfill
        \begin{subfigure}[]{0.3\linewidth}
             \centering
             \includegraphics[width=0.95\linewidth]{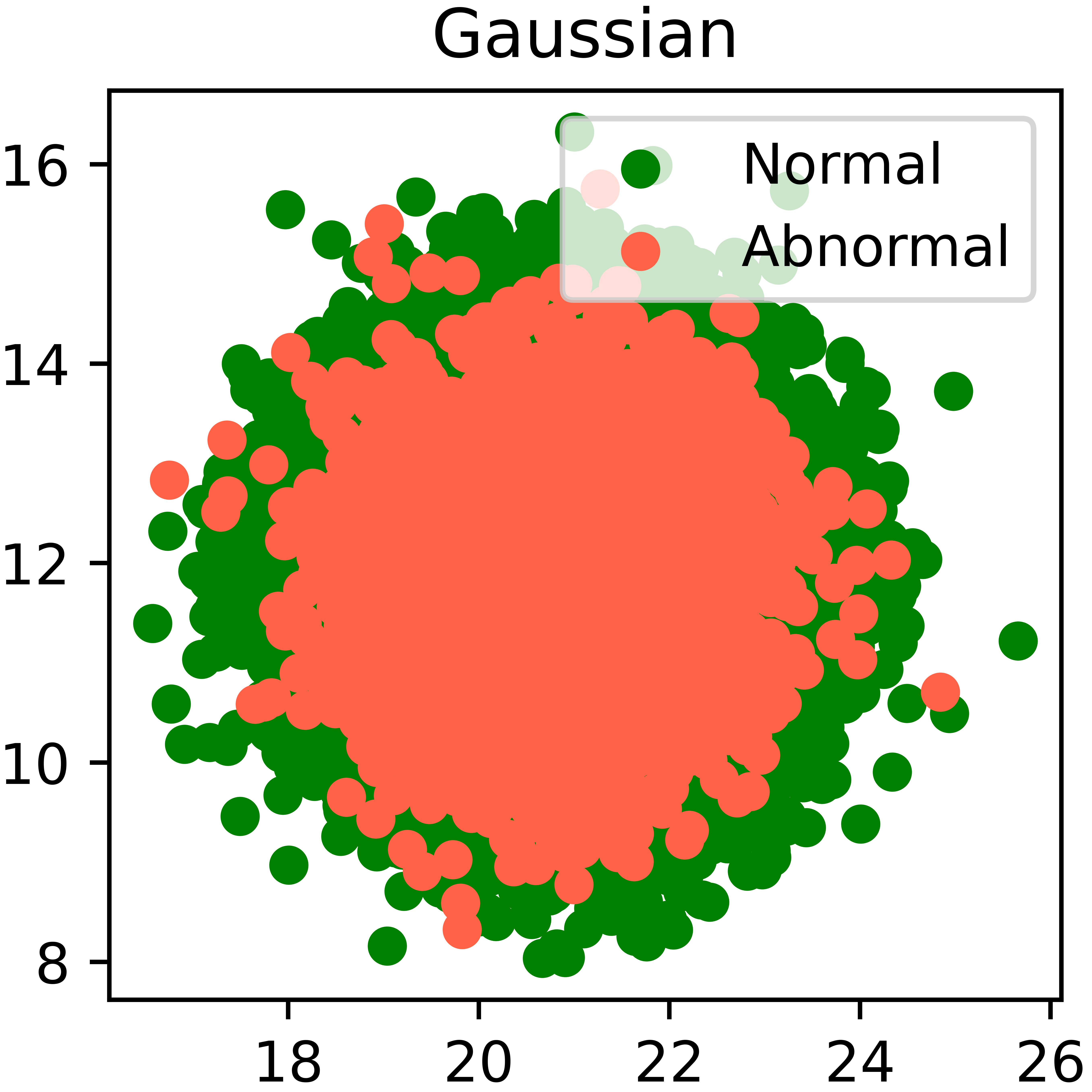}
             \caption{$\tau$=999}  \label{q1_2}
         \end{subfigure}
         \hfill
	\caption{Distribution of both normal and abnormal features at different diffusion step $\tau$.}
	\label{fig:latent_vis}
\end{figure}

To better understand the change of feature distribution \textit{w.r.t.} both normal samples and abnormal samples at different diffusion step $\tau$, here
we use PCA \cite{wold1987principal} for dimension reduction and visualize the data points from MVTec-AD (Wood) as in Figure \ref{fig:latent_vis}.
We can see that anomalies are closer and less distinguishable from normal ones in the near-Gaussian manifold (Figure \ref{q1_1}) than in the original data manifold (Figure \ref{q1_0}).
Therefore, when conducting the denoising generation process of the diffusion model from the near-Gaussian manifold, anomalous samples will have a high probability to be reconstructed into normal samples, which proves the rationality of our motivation to explore the diffusion model as the backbone to mitigate the identity shortcut issue in reconstruction.

\subsection{Evaluation of Hyperparameter Selection}
\label{sec:eval_hyperparam_selection}
Here we study the effectiveness of the proposed strategy that synthesizes anomaly samples to construct a pseudo validation set for hyperparameter selection (Section \ref{pseudo Anomaly Data Synthesizing}).
It is expected that the best parameters selected via the constructed validation set can also lead to the best performance on the real test set.
Specifically, we carry out the hyperparameter selection before inference and consider two hyperparamters: the corruption steps $\tau$ of the latent diffusion model and the number of neighbors $K$ used for feature editing.
We begin by using the model \textit{LafitE w/o F.E.} to identify a $\tau$ that yields the highest detection AUROC on the pseudo validation set.
Next, we fix the selected $\tau$ and repeat this approach to identify the value of $K$ using the LafitE model.

\begin{figure}[htpb]
	\centering
	\begin{subfigure}{0.45\linewidth}
		\includegraphics[width=\linewidth]{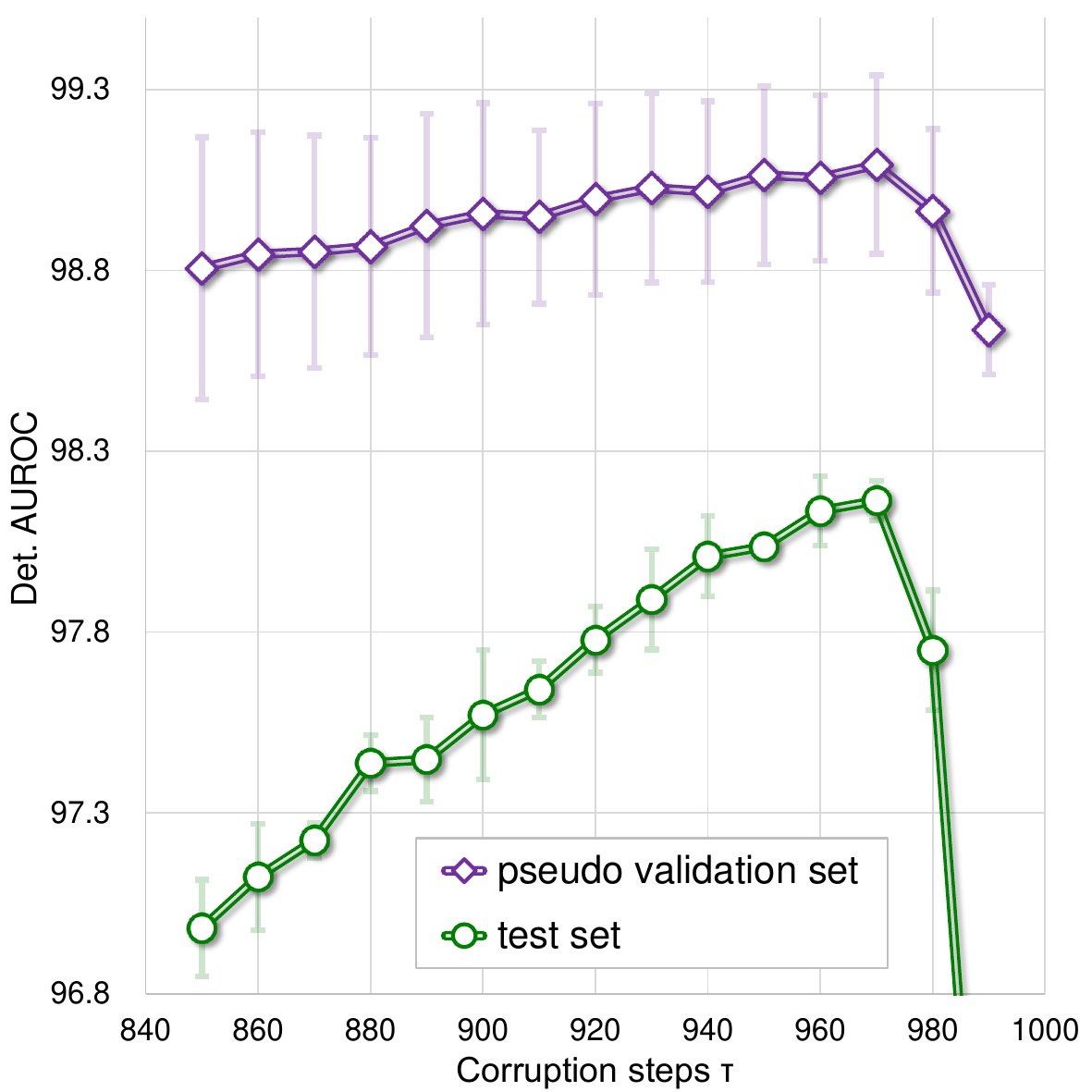}
		\caption{Corruption steps $\tau$}
		\label{hyp-t}
	\end{subfigure}
	\begin{subfigure}{0.45\linewidth}
		\includegraphics[width=\linewidth]{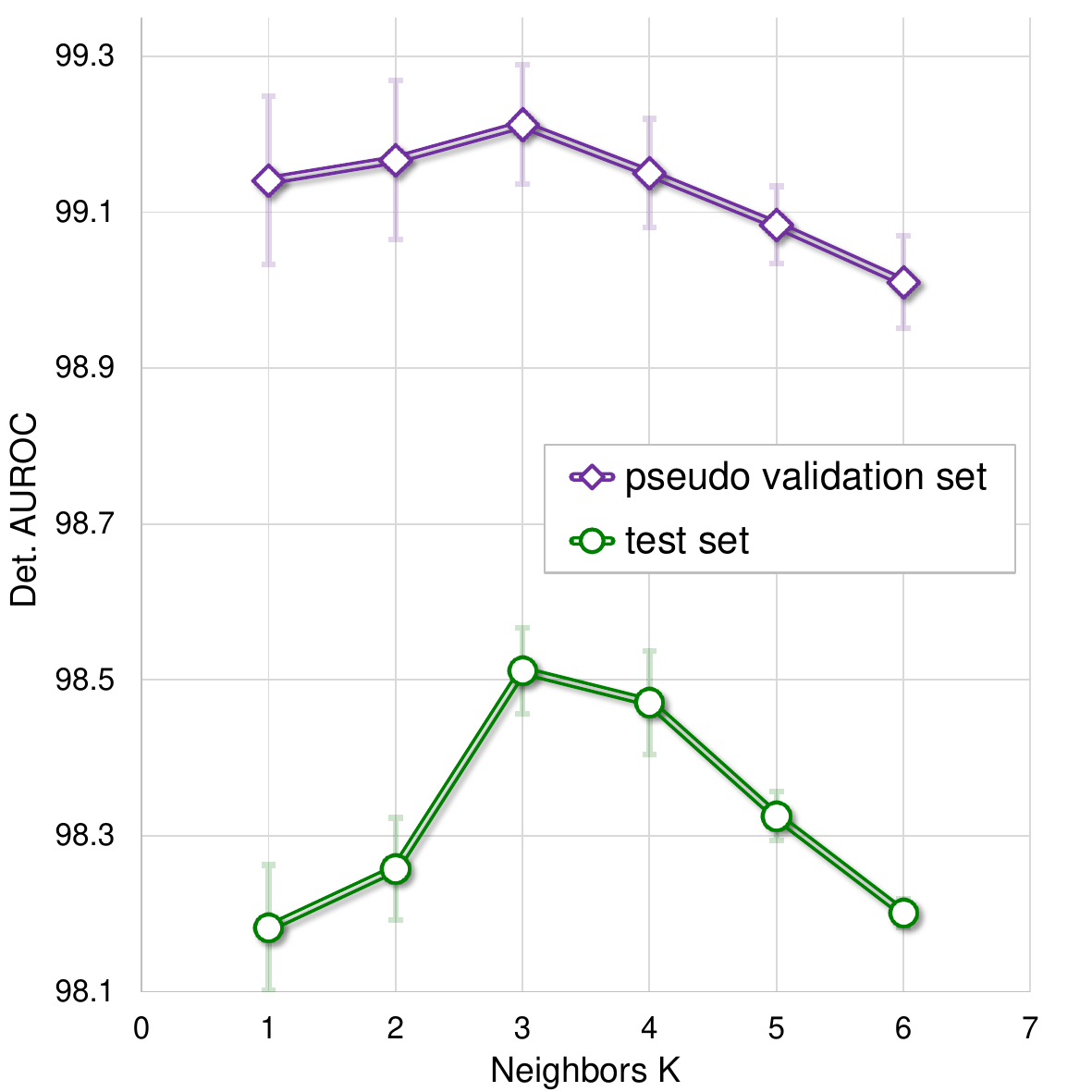}
		\caption{Neighbors $K$}
		\label{hyp-n}
	\end{subfigure}
	\caption{Analysis of the hyperparameter selection strategy.}
	\label{fig:hyparameter}
\end{figure}

Figure \ref{fig:hyparameter} depicts the
detection AUROC obtained from different hyperparameters on both the constructed pseudo validation set and the test set of MVTec-AD \cite{MVTec}. 
Though the gap between the performance curves obtained from the pseudo validation set and the test set suggests the potential improvement of the pseudo validation set,
what is satisfying is that both curves demonstrate similar trends.
More importantly, the hyperparameters along with the best performance on the pseudo validation set also achieve the highest AUROC score on the test set, which ensures that the unsupervised model can perform well in the test phase.


\section{Conclusion}
In this work, we propose an effective framework for unsupervised multi-class anomaly detection: \textbf{LafitE}, \ie, \textbf{La}tent Di\textbf{f}fus\textbf{i}on Model with Fea\textbf{t}ure \textbf{E}diting.
It mainly involves a diffusion model to learn the distribution of normal features and a feature editing strategy to pull the anomalous data toward the normal data manifold, both eliminating the identity shortcut issue in most reconstruction based methods. 
We also highlight the problem of hyperparameter selection in the unsupervised anomaly detection task which is not discussed in prior works.
We emphasize the commonly overlooked challenge of hyperparameter selection in the unsupervised anomaly detection task and propose a strategy to construct a pseudo validation set by synthesizing anomalous samples.
We evaluate the proposed LafitE on MVTec-AD and MPDD datasets.
Experimental results show that our method achieves the SOTA performance on both anomaly detection and anomaly localization.
The qualitative analysis and ablation study demonstrate that the proposed LafitE outperforms the baseline method in solving the identity shortcut issue and our feature editing module can also bring benefits in reducing inaccurate reconstruction.

{\small
\bibliographystyle{ieee_fullname}
\bibliography{reference}

\begin{thebibliography}{10}\itemsep=-1pt

\bibitem{agarwal2005geometric}
Pankaj~K Agarwal et~al.
\newblock Geometric approximation via coresets.
\newblock {\em Combinatorial and computational geometry}, 2005.

\bibitem{Ganomaly}
Samet Akcay, Amir Atapour-Abarghouei, and Toby~P Breckon.
\newblock Ganomaly: Semi-supervised anomaly detection via adversarial training.
\newblock In {\em Asian conference on computer vision}, pages 622--637.
  Springer, 2018.

\bibitem{CycleGANAD}
Christoph Baur, Robert Graf, Benedikt Wiestler, Shadi Albarqouni, and Nassir
  Navab.
\newblock Steganomaly: Inhibiting cyclegan steganography for unsupervised
  anomaly detection in brain mri.
\newblock In {\em International conference on medical image computing and
  computer-assisted intervention}, pages 718--727. Springer, 2020.

\bibitem{AE4MRI}
Christoph Baur, Benedikt Wiestler, Shadi Albarqouni, and Nassir Navab.
\newblock Deep autoencoding models for unsupervised anomaly segmentation in
  brain mr images.
\newblock In {\em International MICCAI brainlesion workshop}, pages 161--169.
  Springer, 2018.

\bibitem{MVTec}
Paul Bergmann, Michael Fauser, David Sattlegger, and Carsten Steger.
\newblock Mvtec ad--a comprehensive real-world dataset for unsupervised anomaly
  detection.
\newblock In {\em Proceedings of the IEEE/CVF conference on computer vision and
  pattern recognition}, pages 9592--9600, 2019.

\bibitem{US}
Paul Bergmann, Michael Fauser, David Sattlegger, and Carsten Steger.
\newblock Uninformed students: Student-teacher anomaly detection with
  discriminative latent embeddings.
\newblock {\em 2020 IEEE/CVF Conference on Computer Vision and Pattern
  Recognition (CVPR)}, pages 4182--4191, 2020.

\bibitem{SPADE}
Niv Cohen and Yedid Hoshen.
\newblock Sub-image anomaly detection with deep pyramid correspondences.
\newblock {\em arXiv preprint arXiv:2005.02357}, 2020.

\bibitem{PaDiM}
Thomas Defard, Aleksandr Setkov, Angelique Loesch, and Romaric Audigier.
\newblock Padim: a patch distribution modeling framework for anomaly detection
  and localization.
\newblock In {\em International Conference on Pattern Recognition}, pages
  475--489. Springer, 2021.

\bibitem{RD4AD}
Hanqiu Deng and Xingyu Li.
\newblock Anomaly detection via reverse distillation from one-class embedding.
\newblock In {\em Proceedings of the IEEE/CVF Conference on Computer Vision and
  Pattern Recognition}, pages 9737--9746, 2022.

\bibitem{DMbeatsGAN}
Prafulla Dhariwal and Alex Nichol.
\newblock Diffusion models beat gans on image synthesis.
\newblock {\em ArXiv}, abs/2105.05233, 2021.

\bibitem{AttributeRestoration}
Ye Fei, Chaoqin Huang, Cao Jinkun, Maosen Li, Ya Zhang, and Cewu Lu.
\newblock Attribute restoration framework for anomaly detection.
\newblock {\em IEEE Transactions on Multimedia}, 2020.

\bibitem{MedicalADSurvey}
Tharindu Fernando, Harshala Gammulle, Simon Denman, Sridha Sridharan, and
  Clinton Fookes.
\newblock Deep learning for medical anomaly detection--a survey.
\newblock {\em ACM Computing Surveys (CSUR)}, 54(7):1--37, 2021.

\bibitem{MemAE}
Dong Gong, Lingqiao Liu, Vuong Le, Budhaditya Saha, Moussa~Reda Mansour, Svetha
  Venkatesh, and Anton van~den Hengel.
\newblock Memorizing normality to detect anomaly: Memory-augmented deep
  autoencoder for unsupervised anomaly detection.
\newblock In {\em Proceedings of the IEEE/CVF International Conference on
  Computer Vision}, pages 1705--1714, 2019.

\bibitem{PretrainedRobustness}
Dan Hendrycks, Kimin Lee, and Mantas Mazeika.
\newblock Using pre-training can improve model robustness and uncertainty.
\newblock In {\em International Conference on Machine Learning}, pages
  2712--2721. PMLR, 2019.

\bibitem{DDPM}
Jonathan Ho, Ajay Jain, and Pieter Abbeel.
\newblock Denoising diffusion probabilistic models.
\newblock {\em Advances in Neural Information Processing Systems},
  33:6840--6851, 2020.

\bibitem{regad}
Chaoqin Huang, Haoyan Guan, Aofan Jiang, Ya Zhang, Michael Spratling, and
  Yan-Feng Wang.
\newblock Registration based few-shot anomaly detection.
\newblock In {\em Computer Vision--ECCV 2022: 17th European Conference, Tel
  Aviv, Israel, October 23--27, 2022, Proceedings, Part XXIV}, pages 303--319.
  Springer, 2022.

\bibitem{MPDD}
Stepan Jezek, Martin Jon{\'a}k, Radim Burget, Pavel Dvorak, and Milos Skotak.
\newblock Deep learning-based defect detection of metal parts: evaluating
  current methods in complex conditions.
\newblock {\em 2021 13th International Congress on Ultra Modern
  Telecommunications and Control Systems and Workshops (ICUMT)}, pages 66--71,
  2021.

\bibitem{SemiOrthog}
Jin-Hwa Kim, Do-Hyeong Kim, Saehoon Yi, and Taehoon Lee.
\newblock Semi-orthogonal embedding for efficient unsupervised anomaly
  segmentation.
\newblock {\em ArXiv}, abs/2105.14737, 2021.

\bibitem{VDM}
Diederik~P. Kingma, Tim Salimans, Ben Poole, and Jonathan Ho.
\newblock Variational diffusion models.
\newblock {\em ArXiv}, abs/2107.00630, 2021.

\bibitem{VAEsystem}
Nejc Kozamernik and Drago Bra{\v{c}}un.
\newblock Visual inspection system for anomaly detection on ktl coatings using
  variational autoencoders.
\newblock {\em Procedia CIRP}, 93:1558--1563, 2020.

\bibitem{ImageNet}
Alex Krizhevsky, Ilya Sutskever, and Geoffrey~E Hinton.
\newblock Imagenet classification with deep convolutional neural networks.
\newblock {\em Communications of the ACM}, 60(6):84--90, 2017.

\bibitem{CutPaste}
Chun-Liang Li, Kihyuk Sohn, Jinsung Yoon, and Tomas Pfister.
\newblock Cutpaste: Self-supervised learning for anomaly detection and
  localization.
\newblock In {\em Proceedings of the IEEE/CVF Conference on Computer Vision and
  Pattern Recognition}, pages 9664--9674, 2021.

\bibitem{SOMAD}
Ning Li, Kaitao Jiang, Zhiheng Ma, Xing Wei, Xiaopeng Hong, and Yihong Gong.
\newblock Anomaly detection via self-organizing map.
\newblock In {\em 2021 IEEE International Conference on Image Processing
  (ICIP)}, pages 974--978. IEEE, 2021.

\bibitem{VisualExplainingVAE}
Wenqian Liu, Runze Li, Meng Zheng, Srikrishna Karanam, Ziyan Wu, Bir Bhanu,
  Richard~J Radke, and Octavia Camps.
\newblock Towards visually explaining variational autoencoders.
\newblock In {\em Proceedings of the IEEE/CVF Conference on Computer Vision and
  Pattern Recognition}, pages 8642--8651, 2020.

\bibitem{loshchilov2017decoupled}
Ilya Loshchilov and Frank Hutter.
\newblock Decoupled weight decay regularization.
\newblock {\em arXiv preprint arXiv:1711.05101}, 2017.

\bibitem{VT-ADL}
Pankaj Mishra, Riccardo Verk, Daniele Fornasier, Claudio Piciarelli, and
  Gian~Luca Foresti.
\newblock Vt-adl: A vision transformer network for image anomaly detection and
  localization.
\newblock In {\em 2021 IEEE 30th International Symposium on Industrial
  Electronics (ISIE)}, pages 01--06. IEEE, 2021.

\bibitem{nichol2021improved}
Alexander~Quinn Nichol and Prafulla Dhariwal.
\newblock Improved denoising diffusion probabilistic models.
\newblock In {\em International Conference on Machine Learning}, pages
  8162--8171. PMLR, 2021.

\bibitem{niu2022memoryBank}
Tongzhi Niu, Bin Li, Weifeng Li, Yuanhong Qiu, and Shuanlong Niu.
\newblock Positive-sample-based surface defect detection using memory-augmented
  adversarial autoencoders.
\newblock {\em IEEE/ASME Transactions on Mechatronics}, 27(1):46--57, 2022.

\bibitem{FastUB}
Walter Hugo~Lopez Pinaya, Mark~S. Graham, Robert Gray, Pedro F.~Da Costa,
  Petru-Daniel Tudosiu, Paul Wright, Yee-Haur Mah, Andrew~D. Mackinnon, James
  Teo, Hans~Rolf J{\"a}ger, David~John Werring, Geraint Rees, Parashkev Nachev,
  S{\'e}bastien Ourselin, and Manuel~Jorge Cardoso.
\newblock Fast unsupervised brain anomaly detection and segmentation with
  diffusion models.
\newblock In {\em MICCAI}, 2022.

\bibitem{InpaintingTransformer}
Jonathan Pirnay and Keng Chai.
\newblock Inpainting transformer for anomaly detection.
\newblock In {\em International Conference on Image Analysis and Processing},
  pages 394--406. Springer, 2022.

\bibitem{VideoADSurvey}
Bharathkumar Ramachandra, Michael Jones, and Ranga~Raju Vatsavai.
\newblock A survey of single-scene video anomaly detection.
\newblock {\em IEEE transactions on pattern analysis and machine intelligence},
  2020.

\bibitem{PANDA}
Tal Reiss, Niv Cohen, Liron Bergman, and Yedid Hoshen.
\newblock Panda: Adapting pretrained features for anomaly detection and
  segmentation.
\newblock In {\em Proceedings of the IEEE/CVF Conference on Computer Vision and
  Pattern Recognition}, pages 2806--2814, 2021.

\bibitem{GaussianFT}
Oliver Rippel, Arnav Chavan, Chucai Lei, and Dorit Merhof.
\newblock Transfer learning gaussian anomaly detection by fine-tuning
  representations.
\newblock {\em Proceedings of the 2nd International Conference on Image
  Processing and Vision Engineering}, 2021.

\bibitem{PretrainedModeling}
Oliver Rippel, Patrick Mertens, and Dorit Merhof.
\newblock Modeling the distribution of normal data in pre-trained deep features
  for anomaly detection.
\newblock In {\em 2020 25th International Conference on Pattern Recognition
  (ICPR)}, pages 6726--6733. IEEE, 2021.

\bibitem{LDM}
Robin Rombach, A. Blattmann, Dominik Lorenz, Patrick Esser, and Bj{\"o}rn
  Ommer.
\newblock High-resolution image synthesis with latent diffusion models.
\newblock {\em 2022 IEEE/CVF Conference on Computer Vision and Pattern
  Recognition (CVPR)}, pages 10674--10685, 2022.

\bibitem{U-net}
Olaf Ronneberger, Philipp Fischer, and Thomas Brox.
\newblock U-net: Convolutional networks for biomedical image segmentation.
\newblock In {\em International Conference on Medical image computing and
  computer-assisted intervention}, pages 234--241. Springer, 2015.

\bibitem{PatchCore}
Karsten Roth, Latha Pemula, Joaquin Zepeda, Bernhard Sch{\"o}lkopf, Thomas
  Brox, and Peter Gehler.
\newblock Towards total recall in industrial anomaly detection.
\newblock In {\em Proceedings of the IEEE/CVF Conference on Computer Vision and
  Pattern Recognition}, pages 14318--14328, 2022.

\bibitem{DeepSVDD}
Lukas Ruff, Robert Vandermeulen, Nico Goernitz, Lucas Deecke, Shoaib~Ahmed
  Siddiqui, Alexander Binder, Emmanuel M{\"u}ller, and Marius Kloft.
\newblock Deep one-class classification.
\newblock In {\em International conference on machine learning}, pages
  4393--4402. PMLR, 2018.

\bibitem{SemiSVDD}
Lukas Ruff, Robert~A Vandermeulen, Nico G{\"o}rnitz, Alexander Binder, Emmanuel
  M{\"u}ller, Klaus-Robert M{\"u}ller, and Marius Kloft.
\newblock Deep semi-supervised anomaly detection.
\newblock {\em arXiv preprint arXiv:1906.02694}, 2019.

\bibitem{ALOCC}
Mohammad Sabokrou, Mohammad Khalooei, Mahmood Fathy, and Ehsan Adeli.
\newblock Adversarially learned one-class classifier for novelty detection.
\newblock In {\em Proceedings of the IEEE conference on computer vision and
  pattern recognition}, pages 3379--3388, 2018.

\bibitem{fisrtANAE}
Mayu Sakurada and Takehisa Yairi.
\newblock Anomaly detection using autoencoders with nonlinear dimensionality
  reduction.
\newblock In {\em Proceedings of the MLSDA 2014 2nd workshop on machine
  learning for sensory data analysis}, pages 4--11, 2014.

\bibitem{MKD}
Mohammadreza Salehi, Niousha Sadjadi, Soroosh Baselizadeh, Mohammad~Hossein
  Rohban, and Hamid~R. Rabiee.
\newblock Multiresolution knowledge distillation for anomaly detection.
\newblock {\em 2021 IEEE/CVF Conference on Computer Vision and Pattern
  Recognition (CVPR)}, pages 14897--14907, 2021.

\bibitem{fAnoGAN}
Thomas Schlegl, Philipp Seeb{\"o}ck, Sebastian~M Waldstein, Georg Langs, and
  Ursula Schmidt-Erfurth.
\newblock f-anogan: Fast unsupervised anomaly detection with generative
  adversarial networks.
\newblock {\em Medical image analysis}, 54:30--44, 2019.

\bibitem{OC-SVM}
Bernhard Sch{\"o}lkopf, John~C Platt, John Shawe-Taylor, Alex~J Smola, and
  Robert~C Williamson.
\newblock Estimating the support of a high-dimensional distribution.
\newblock {\em Neural computation}, 13(7):1443--1471, 2001.

\bibitem{SVM4AD}
Bernhard Sch{\"o}lkopf, Robert~C Williamson, Alex Smola, John Shawe-Taylor, and
  John Platt.
\newblock Support vector method for novelty detection.
\newblock {\em Advances in neural information processing systems}, 12, 1999.

\bibitem{DFR}
Yong Shi, Jie Yang, and Zhiquan Qi.
\newblock Unsupervised anomaly segmentation via deep feature reconstruction.
\newblock {\em Neurocomputing}, 424:9--22, 2021.

\bibitem{DDIM}
Jiaming Song, Chenlin Meng, and Stefano Ermon.
\newblock Denoising diffusion implicit models.
\newblock {\em arXiv preprint arXiv:2010.02502}, 2020.

\bibitem{tan2021trustmae}
Daniel~Stanley Tan, Yi-Chun Chen, Trista Pei-Chun Chen, and Wei-Chao Chen.
\newblock Trustmae: A noise-resilient defect classification framework using
  memory-augmented auto-encoders with trust regions.
\newblock In {\em Proceedings of the IEEE/CVF winter conference on applications
  of computer vision}, pages 276--285, 2021.

\bibitem{tan2019efficientnet}
Mingxing Tan and Quoc Le.
\newblock Efficientnet: Rethinking model scaling for convolutional neural
  networks.
\newblock In {\em International conference on machine learning}, pages
  6105--6114. PMLR, 2019.

\bibitem{Tao2022DeepLF}
Xian Tao, Xinyi Gong, Xin~Yu Zhang, Shaohua Yan, and Chandranath Adak.
\newblock Deep learning for unsupervised anomaly localization in industrial
  images: A survey.
\newblock {\em IEEE Transactions on Instrumentation and Measurement}, 71:1--21,
  2022.

\bibitem{SVDD}
David~MJ Tax and Robert~PW Duin.
\newblock Support vector data description.
\newblock {\em Machine learning}, 54(1):45--66, 2004.

\bibitem{vaswani2017attention}
Ashish Vaswani, Noam Shazeer, Niki Parmar, Jakob Uszkoreit, Llion Jones,
  Aidan~N Gomez, {\L}ukasz Kaiser, and Illia Polosukhin.
\newblock Attention is all you need.
\newblock {\em Advances in neural information processing systems}, 30, 2017.

\bibitem{AttenVAE}
Shashanka Venkataramanan, Kuan-Chuan Peng, Rajat~Vikram Singh, and Abhijit
  Mahalanobis.
\newblock Attention guided anomaly localization in images.
\newblock In {\em European Conference on Computer Vision}, pages 485--503.
  Springer, 2020.

\bibitem{wold1987principal}
Svante Wold, Kim Esbensen, and Paul Geladi.
\newblock Principal component analysis.
\newblock {\em Chemometrics and intelligent laboratory systems}, 2(1-3):37--52,
  1987.

\bibitem{metaformer}
Jhih-Ciang Wu, Ding-Jie Chen, Chiou-Shann Fuh, and Tyng-Luh Liu.
\newblock Learning unsupervised metaformer for anomaly detection.
\newblock In {\em Proceedings of the IEEE/CVF International Conference on
  Computer Vision}, pages 4369--4378, 2021.

\bibitem{AnoDDPMAD}
Julian Wyatt, Adam Leach, Sebastian~M. Schmon, and Chris~G. Willcocks.
\newblock Anoddpm: Anomaly detection with denoising diffusion probabilistic
  models using simplex noise.
\newblock {\em 2022 IEEE/CVF Conference on Computer Vision and Pattern
  Recognition Workshops (CVPRW)}, pages 649--655, 2022.

\bibitem{yang2019clustering}
Hua Yang, Yifan Chen, Kaiyou Song, and Zhouping Yin.
\newblock Multiscale feature-clustering-based fully convolutional autoencoder
  for fast accurate visual inspection of texture surface defects.
\newblock {\em IEEE Transactions on Automation Science and Engineering},
  16(3):1450--1467, 2019.

\bibitem{PatchSVDD}
Jihun Yi and Sungroh Yoon.
\newblock Patch svdd: Patch-level svdd for anomaly detection and segmentation.
\newblock In {\em Proceedings of the Asian Conference on Computer Vision},
  2020.

\bibitem{UniAD}
Zhiyuan You, Lei Cui, Yujun Shen, Kai Yang, Xin Lu, Yu Zheng, and Xinyi Le.
\newblock A unified model for multi-class anomaly detection.
\newblock {\em Advances in Neural Information Processing Systems}, 2022.

\bibitem{yu2021fastflow}
Jiawei Yu, Ye Zheng, Xiang Wang, Wei Li, Yushuang Wu, Rui Zhao, and Liwei Wu.
\newblock Fastflow: Unsupervised anomaly detection and localization via 2d
  normalizing flows.
\newblock {\em arXiv preprint arXiv:2111.07677}, 2021.

\bibitem{DRAEM}
Vitjan Zavrtanik, Matej Kristan, and Danijel Sko{\v{c}}aj.
\newblock Draem-a discriminatively trained reconstruction embedding for surface
  anomaly detection.
\newblock In {\em Proceedings of the IEEE/CVF International Conference on
  Computer Vision}, pages 8330--8339, 2021.

\bibitem{RIAD}
Vitjan Zavrtanik, Matej Kristan, and Danijel Sko{\v{c}}aj.
\newblock Reconstruction by inpainting for visual anomaly detection.
\newblock {\em Pattern Recognition}, 112:107706, 2021.

\bibitem{zhao2017spatio}
Yiru Zhao, Bing Deng, Chen Shen, Yao Liu, Hongtao Lu, and Xian-Sheng Hua.
\newblock Spatio-temporal autoencoder for video anomaly detection.
\newblock In {\em Proceedings of the 25th ACM international conference on
  Multimedia}, pages 1933--1941, 2017.

\bibitem{GMM4AD}
Bo Zong, Qi Song, Martin~Renqiang Min, Wei Cheng, Cristian Lumezanu, Daeki Cho,
  and Haifeng Chen.
\newblock Deep autoencoding gaussian mixture model for unsupervised anomaly
  detection.
\newblock In {\em International conference on learning representations}, 2018.

\bibitem{zou2022spot}
Yang Zou, Jongheon Jeong, Latha Pemula, Dongqing Zhang, and Onkar Dabeer.
\newblock Spot-the-difference self-supervised pre-training for anomaly
  detection and segmentation.
\newblock In {\em Computer Vision--ECCV 2022: 17th European Conference, Tel
  Aviv, Israel, October 23--27, 2022, Proceedings, Part XXX}, pages 392--408.
  Springer, 2022.

\end{thebibliography}
}

\end{document}


\title{LafitE: Latent Diffusion Model with Feature Editing for \\Unsupervised Multi-class Anomaly Detection\\Supplementary Material}

\maketitle
\thispagestyle{empty}
\appendix



\section{Additional Quantitative Performance}
We augment this section with experiments on the AUROC metric in the separate case, as well as the AUPRO\cite{US} metric in the unified case. We present metrics for both anomaly detection and anomaly localization on each category in Table \ref{tab:Separate-MVtec} and Table \ref{tab:Separate-MPDD}, respectively. Our results reveal that, on the MVTec-AD\cite{MVTec} dataset, LafitE exhibits comparable performance to the baselines, while on the MPDD\cite{MPDD} dataset, LafitE outperforms all the baselines, thus showcasing the robustness of our proposed method.

We conduct experiments for UniAD\cite{UniAD} and our LafitE with the AUPRO\cite{US} metric in unified case on the MVTec-AD\cite{MVTec} and MPDD\cite{MPDD} dataset, respectively. The results are shown in Table \ref{tab:PRO-MvTec-United} and Table \ref{tab:PRO-MPDD-United}.
Table \ref{tab:PRO-MvTec-United} demonstrates that LafitE achieves higher AUPRO metrics than UniAD for most categories, outperforming UniAD by a significant margin of \textbf{3.1\%}. 
This clearly indicates the superior performance of LafitE compared to UniAD. 
Futhermore, LafitE without feature editing reduces the AUPRO by 0.6\% on average, again validating the effectiveness of feature editing in our method.
We also present AUPRO metrics on the MPDD dataset in Table \ref{tab:PRO-MPDD-United}. 
For the PRO metrics that are indicative of anomaly localization capability, LafitE comprehensively outperformed UniAD with an average performance of \textbf{12.4\%} higher on MPDD.

\begin{table}[htpb]
    \centering
    \caption{AUPRO performance of all categories on MVTec-AD \cite{MVTec}.}
    \resizebox{0.48\textwidth}{!}{
    \begin{tabular}{c|ccc}
    \toprule
        \multirow{2}*{Category} & \multicolumn{3}{c}{Methods} \\
        \cmidrule{2-4}
                                & UniAD\cite{UniAD} & LafitE w/o F.E. & LafitE (ours) \\
         \midrule
          Bottle        & 93.7 & 95.4 $\pm$ \small{0.14} & \textbf{95.7} $\pm$ \small{0.05} \\
          Cable         & 86.8 & 88.9 $\pm$ \small{0.43} & \textbf{91.6} $\pm$ \small{0.14} \\
          Capsule       & 89.8 & 92.9 $\pm$ \small{0.22} & \textbf{93.3} $\pm$ \small{0.16} \\
          Hazelnut      & 93.3 & 94.6 $\pm$ \small{0.34} & \textbf{94.8} $\pm$ \small{0.28} \\
          Metal Nut     & 80.0 & 90.3 $\pm$ \small{0.27} & \textbf{91.8} $\pm$ \small{0.09} \\
          Pill          & 95.0 & 96.4 $\pm$ \small{0.10} & \textbf{96.6} $\pm$ \small{0.07} \\
          Screw         & 93.4 & 97.0 $\pm$ \small{0.16} & \textbf{97.1} $\pm$ \small{0.03} \\
          Toothbrush    & 87.7 & \textbf{91.2} $\pm$ \small{0.10} & 90.9 $\pm$ \small{0.13} \\
          Transistor    & \textbf{95.2} & 94.8 $\pm$ \small{0.12} & 94.7 $\pm$ \small{0.10} \\
          Zipper        & 90.2 & \textbf{95.2} $\pm$ \small{0.21} & 94.6 $\pm$ \small{0.06} \\
          Carpet        & 94.7 & 94.4 $\pm$ \small{0.13} & \textbf{95.2} $\pm$ \small{0.06} \\
          Grid          & 90.7 & 95.5 $\pm$ \small{0.20} & \textbf{95.7} $\pm$ \small{0.15} \\
          Leather       & 97.3 & 97.1 $\pm$ \small{0.11} & \textbf{98.0} $\pm$ \small{0.02} \\
          Tile          & 82.1 & 83.1 $\pm$ \small{0.74} & \textbf{84.9} $\pm$ \small{0.56} \\
          Wood          & 88.4 & 88.9 $\pm$ \small{0.18} & \textbf{90.2} $\pm$ \small{0.25} \\
          \midrule
          Average       & 90.6 & 93.1 $\pm$ \small{0.06} & \textbf{93.7} $\pm$ \small{0.07} \\
        \bottomrule     
    \end{tabular}
    }
    \label{tab:PRO-MvTec-United}
\end{table}

\begin{table}[htpb]
    \centering
    \caption{AUPRO performance of all categories on MPDD \cite{MPDD}.}
    \resizebox{0.48\textwidth}{!}{
    \begin{tabular}{c|ccc}
    \toprule
        \multirow{2}*{Category}            & \multicolumn{3}{c}{Methods} \\
        \cmidrule{2-4}
         & UniAD\cite{UniAD} & LafitE w/o F.E. & LafitE (ours) \\
         \midrule
          Bracket Black & 84.1 & 94.4 $\pm$ \small{0.04} & \textbf{98.0} $\pm$ \small{0.02} \\
          Bracket Brown & 86.6 & 89.6 $\pm$ \small{0.16} & \textbf{96.6} $\pm$ \small{0.03} \\
          Bracket White & 81.4 & 88.5 $\pm$ \small{0.22} & \textbf{92.0} $\pm$ \small{0.10} \\
          Connector     & 93.1 & 95.7 $\pm$ \small{0.06} & \textbf{96.9} $\pm$ \small{0.09} \\
          Metal Plate   & 82.6 & 93.5 $\pm$ \small{0.05} & \textbf{94.4} $\pm$ \small{0.06} \\
          Tubes         & 72.8 & 92.1 $\pm$ \small{0.21} & \textbf{96.8} $\pm$ \small{0.08} \\
          \midrule
          Average       & 83.4 & 92.3 $\pm$ \small{0.04} & \textbf{95.8} $\pm$ \small{0.03} \\
        \bottomrule     
    \end{tabular}
    }
    \label{tab:PRO-MPDD-United}
\end{table}

\begin{table*}[!t]
    \centering
    \setlength{\tabcolsep}{1mm}
    \caption{AUROC scores of anomaly \textbf{detection} and \textbf{localization} (shown as Det. / Loc.) on MVTec-AD \cite{MVTec} in \textbf{separate} case.
    }
    \resizebox{0.99\textwidth}{!}{
    \begin{tabular}{c|ccccccc|c}
    \toprule
         \textbf{Category} 
         & \textbf{PatchSVDD} \cite{PatchSVDD} 
         & \textbf{PaDiM} \cite{PaDiM} 
         & \textbf{DRAEM} \cite{DRAEM} 
         & \textbf{RevDistill} \cite{RD4AD} 
         & \textbf{PatchCore} \cite{PatchCore} 
         & \textbf{FastFlow} \cite{yu2021fastflow} 
         & \textbf{UniAD} \cite{UniAD} 
         & \textbf{LafitE (ours)} \\
         \midrule
          Bottle 
          & 98.6 / 98.1
          & 99.9 / 98.2
          & 99.2 / 99.1
          & 100 / 98.7
          & 100 / 98.6
          & 100 / 97.7
          & 100 / 98.1
          & 100 / 98.6
          \\
           Cable 
           & 90.3 / 96.8
           & 92.7 / 96.7
           & 91.8 / 94.7
           & 95.0 / 97.4
           & 99.5 / 98.4
           & 100 / 98.4
           & 97.6 / 96.8
           & 99.0 / 98.3
          \\
           Capsule 
           & 76.7 / 95.8
           & 91.3 / 98.6
           & 98.5 / 94.3
           & 96.3 / 98.7
           & 98.1 / 98.8
           & 100 / 99.1
           & 85.3 / 97.9
           & 97.7 / 98.9
          \\
           Hazelnut 
           & 92.0 / 97.5
           & 92.0 / 98.1
           & 100 / 99.7
           & 99.9 / 98.9
           & 100 / 98.7
           & 100 / 99.1
           & 99.9 / 98.8
           & 100 / 98.7
          \\
           Metal Nut 
           & 81.3 / 98.0
           & 98.7 / 97.3
           & 98.7 / 99.5
           & 100 / 97.3
           & 100 / 98.4
           & 100 / 98.5
           & 99.0 / 95.7
           & 99.7 / 96.6
          \\
           Pill  
           & 86.1 / 95.1
           & 93.3 / 95.7
           & 98.9 / 97.6
           & 96.6 / 98.2
           & 96.6 / 97.4
           & 99.4 / 99.2
           & 88.3 / 95.1
           & 97.5 / 97.2
          \\
           Screw 
           & 81.3 / 95.7
           & 85.8 / 98.4
           & 93.9 / 97.6
           & 97.0 / 99.6
           & 98.1 / 99.4
           & 97.8 / 99.4
           & 91.9 / 97.4
           & 94.1 / 99.5
          \\
          Toothbrush 
          & 100 / 98.1
          & 96.1 / 98.8
          & 100 / 98.1
          & 99.5 / 99.1
          & 100 / 98.7
          & 94.4 / 98.9
          & 95.0 / 97.8 
          & 91.7 / 98.8
          \\
          Transistor 
          & 91.5 / 97.0
          & 97.4 / 97.6
          & 93.1 / 90.9
          & 96.7 / 92.5
          & 100 / 96.3
          & 99.8 / 97.3
          & 100 / 98.7
          & 99.9 / 98.4
          \\
          Zipper 
          & 97.9 / 95.1
          & 90.3 / 98.4
          & 100 / 98.8
          & 98.5 / 98.2
          & 99.4 / 98.8
          & 99.5 / 98.7
          & 96.7 / 96.0 
          & 99.1 / 98.1
          \\
          Carpet 
          & 92.9 / 92.6
          & 99.8 / 99.0
          & 97.0 / 95.5
          & 98.9 / 98.9
          & 98.7 / 99.0
          & 100 / 99.4
          & 99.9 / 98.0
          & 99.9 / 98.7
          \\
          grid 
          & 94.6 / 96.2
          & 96.7 / 97.1
          & 99.9 / 99.7
          & 100 / 99.3
          & 98.2 / 98.7
          & 99.7 / 98.3
          & 98.5 / 94.6
          & 100 / 98.5
          \\
          Leather 
          & 90.9 / 97.4
          & 100 / 99.0
          & 100 / 98.6
          & 100 / 99.4
          & 100 / 99.3
          & 100 / 99.5
          & 100 / 98.3
          & 100 / 99.2
          \\
          Tile
          & 97.8 / 91.4
          & 98.1 / 94.1
          & 99.6 / 99.2
          & 99.3 / 95.6
          & 98.7 / 95.6
          & 100 / 96.3
          & 99.0 / 91.8
          & 99.9 / 92.4
          \\
          Wood 
          & 96.5 / 90.8
          & 99.2 / 94.1
          & 99.1 / 96.4
          & 99.2 / 95.3
          & 99.2 / 95.0
          & 100 / 97.0
          & 97.9 / 93.4
          & 98.9 / 94.4
          \\
          \midrule
          \rowcolor{gray!10}
          \textbf{Average}
          & 92.1 / 95.7
          & 95.5 / 97.4
          & 98.0 / 97.3
          & 98.5 / 97.8
          & 99.1 / 98.1
          & 99.4 / 98.5
          & 96.6 / 96.6
          & 98.5 / 97.8
          \\
        \bottomrule     
    \end{tabular}
    }
    \label{tab:Separate-MVtec}
\end{table*}

\begin{table*}[!t]
    \centering
    \setlength{\tabcolsep}{1mm}
    \caption{
    AUROC scores of anomaly \textbf{detection} and \textbf{localization} (shown as Det. / Loc.) on MPDD \cite{MPDD} in \textbf{separate} case.
    }
    \resizebox{0.99\textwidth}{!}{
    \begin{tabular}{c|ccccccc|c}
    \toprule
         \textbf{Category} 
         & \textbf{PatchSVDD} \cite{PatchSVDD} 
         & \textbf{PaDiM} \cite{PaDiM} 
         & \textbf{DRAEM} \cite{DRAEM} 
         & \textbf{RevDistill} \cite{RD4AD} 
         & \textbf{PatchCore} \cite{PatchCore} 
         & \textbf{FastFlow} \cite{yu2021fastflow} 
         & \textbf{UniAD} \cite{UniAD} 
         & \textbf{LafitE (ours)} \\
         \midrule
          Bracket Black  
          & 88.3 / 44.7
          & 75.6 / 94.2
          & 80.5 / 97.7
          & 78.2 / 97.2
          & 81.9 / 98.4
          & 89.6 / 96.3
          & 87.2 / 94.0
          & 98.6 / 99.3
          \\
          Bracket Brown  
          & 98.8 / 67.3
          & 85.4 / 92.4
          & 88.1 / 59.7
          & 91.0 / 96.9
          & 78.4 / 91.5
          & 97.4 / 95.3
          & 93.0 / 98.3
          & 96.9 / 99.4
          \\
          Bracket White  
          & 95.9 / 53.0
          & 82.2 / 98.1
          & 92.3 / 98.8
          & 85.8 / 99.2
          & 76.0 / 97.4
          & 88.2 / 98.1
          & 53.0 / 93.8
          & 91.4 / 98.2
          \\
          Connector  
          & 96.7 / 83.1
          & 91.7 / 97.9
          & 99.5 / 94.7
          & 99.0 / 99.4
          & 96.7 / 95.0
          & 95.0 / 97.2
          & 88.6 / 96.6
          & 96.7 / 99.1
          \\
          Metal Plate  
          & 98.4 / 95.2
          & 56.3 / 92.9
          & 100 / 97.5
          & 99.9 / 99.0
          & 100 / 96.6
          & 100 / 98.9
          & 57.2 / 87.7
          & 100 / 98.8
          \\
          Tubes  
          & 95.2 / 56.9
          & 57.5 / 93.9
          & 93.8 / 96.6
          & 94.2 / 99.1
          & 59.7 / 95.1
          & 98.1 / 99.4
          & 66.1 / 91.7
          & 95.7 / 99.2
          \\

          \midrule
          \rowcolor{gray!10}
          \textbf{Average}
          & 95.6 / 66.7
          & 74.8 / 94.9
          & 92.4 / 90.8
          & 91.4 / 98.5
          & 82.1 / 95.7
          & 94.7 / 97.6
          & 74.2 / 93.7
          & 96.5 / 99.0
          \\
        \bottomrule     
    \end{tabular}
    }
    \label{tab:Separate-MPDD}
\end{table*}

\section{Details of Hyperparameter Selection}

In the main paper we propose a straightforward method to synthesize anomaly samples for facilitating hyperparameter selection. 
To implement our approach, we begin by generating a randomized mask, followed by a randomized shuffling of the masked areas in each image within a batch, resulting in a novel set of pseudo-abnormal data, while still retaining a fraction of normal samples in the synthesized validation set.  
We set the batch size to 16 for anomalous data synthesis, ensuring an appropriate number of normal samples in the validation set.  

Our proposed hyperparameter selection method was employed in the experiments detailed in our main paper to facilitate the selection of optimal hyperparameters for inference using on the MVTec-AD\cite{MVTec} and MPDD\cite{MPDD} datasets. 
Specifically, we select $\tau=970$ and $K=3$ for MVTec-AD, and $\tau=860$ and $K=3$ for MPDD which were the most suitable ones.



\section{Core Set Sampling for Feature Editing}

In feature editing, a core set is extracted from the entire memory bank via a sampling process. This process employs a greedy search algorithm\cite{agarwal2005geometric} that ensures the resulting samples in the core set are distributed as uniformly as possible within the latent space. Details regarding the sampling algorithm can be found in Algorithm \ref{alg:core_set}.

\begin{algorithm}
\SetAlgoLined
\KwInput{Whole Memory bank $\mathcal{M}$, keep rate $\textit{r}$.}
\KwOutput{Core Set of the Memory bank $\mathcal{M}_C$.}
\textbf{Algorithm:}\\
$\mathcal{M}_C \leftarrow \{\}$ \\
$n_c \leftarrow \lfloor |\mathcal{M}| \times \textit{r} \rfloor $ \\
\For{$i\in[0, ..., n_c - 1]$}{
	$m_i \leftarrow \underset{m\in\mathcal{M}-\mathcal{M}_C}{\argmax}\underset{m^{'}\in\mathcal{M}_C}{\min} \left\Vert m - m^{'} \right\Vert_2$ \\
	$\mathcal{M}_C \leftarrow \mathcal{M}_C \cup \{m_i\}$\\
}
\caption{Core Set Sampling.}
\label{alg:core_set}
\end{algorithm}

\section{Visualization Results}

Figure \ref{fig:visualization_MvTec} provides additional visualization results for categories not presented in the main paper on the MVTec-AD\cite{MVTec} dataset. 
In addition to the visualization of reconstruction, we show two anomaly localization heatmaps. One is normalized on a single category, the other is normalized on a single image, which is more widely used for visual presentation.
While the main paper presented visualization results using single category normalization, we refrain from utilizing single image normalization for anomaly heatmap generation, as it tends to produce high anomaly values for normal images as well, which is not a reasonable approach in practical applications.

From Figure \ref{fig:visualization_MvTec}, we can observe that compared to UniAD\cite{UniAD},
our method is more accurate in locating anomalies in most of the categories. 
This is attributable to either our method's superior reconstruction capabilities (\eg, screw, carpet, grid, bottle, metalnut), or the ability to retain more detailed original features in non-abnormal regions (\eg, zipper, tile, transistor, capsule).
Our method outperforms UniAD in solving both false negative and false positive issues, again verifying the advantage of LafitE.

The results of the visualization comparison between our method and UniAD on the MPDD\cite{MPDD} dataset are shown in Figure \ref{fig:visualization_MPDD}.
It is obvious that our method has better generation effect on more complex datasets like MPDD, \ie., the anomalies are removed while the original image details are better preserved.
This enables our method to have a superior effect on anomaly localization, which is especially obvious on metal plates and tubes.

\begin{figure*}[t]
     \centering
     \begin{subfigure}{\linewidth}
         \centering
         \includegraphics[width=0.12\linewidth]{Figure_supp/MvTeccase/bottle/ori.png}\hspace{-0.07cm}
         \includegraphics[width=0.12\linewidth]{Figure_supp/MvTeccase/bottle/gt.png}\hspace{-0.07cm}
         \includegraphics[width=0.12\linewidth]{Figure_supp/MvTeccase/bottle/UniAD_recon.png}\hspace{-0.07cm}
         \includegraphics[width=0.12\linewidth]{Figure_supp/MvTeccase/bottle/ddpm_membank_guassfilter_recon.png}
         \includegraphics[width=0.12\linewidth]{Figure_supp/MvTeccase/bottle/UniAD_heatmap.png}\hspace{-0.07cm}
         \includegraphics[width=0.12\linewidth]{Figure_supp/MvTeccase/bottle/ddpm_membank_guassfilter_heatmap.png}\hspace{-0.07cm}
         \includegraphics[width=0.12\linewidth]{Figure_supp/MvTeccase/bottle/UniAD_heatmapself.png}\hspace{-0.07cm}
         \includegraphics[width=0.12\linewidth]{Figure_supp/MvTeccase/bottle/ddpm_membank_guassfilter_heatmapself.png}
     \end{subfigure}
     \begin{subfigure}{\linewidth}
         \centering
         \includegraphics[width=0.12\linewidth]{Figure_supp/MvTeccase/capsule/ori.png}\hspace{-0.07cm}
         \includegraphics[width=0.12\linewidth]{Figure_supp/MvTeccase/capsule/gt.png}\hspace{-0.07cm}
         \includegraphics[width=0.12\linewidth]{Figure_supp/MvTeccase/capsule/UniAD_recon.png}\hspace{-0.07cm}
         \includegraphics[width=0.12\linewidth]{Figure_supp/MvTeccase/capsule/ddpm_membank_guassfilter_recon.png}
         \includegraphics[width=0.12\linewidth]{Figure_supp/MvTeccase/capsule/UniAD_heatmap.png}\hspace{-0.07cm}
         \includegraphics[width=0.12\linewidth]{Figure_supp/MvTeccase/capsule/ddpm_membank_guassfilter_heatmap.png}\hspace{-0.07cm}
         \includegraphics[width=0.12\linewidth]{Figure_supp/MvTeccase/capsule/UniAD_heatmapself.png}\hspace{-0.07cm}
         \includegraphics[width=0.12\linewidth]{Figure_supp/MvTeccase/capsule/ddpm_membank_guassfilter_heatmapself.png}
     \end{subfigure}
     \begin{subfigure}{\linewidth}
         \centering
         \includegraphics[width=0.12\linewidth]{Figure_supp/MvTeccase/carpet/ori.png}\hspace{-0.07cm}
         \includegraphics[width=0.12\linewidth]{Figure_supp/MvTeccase/carpet/gt.png}\hspace{-0.07cm}
         \includegraphics[width=0.12\linewidth]{Figure_supp/MvTeccase/carpet/UniAD_recon.png}\hspace{-0.07cm}
         \includegraphics[width=0.12\linewidth]{Figure_supp/MvTeccase/carpet/ddpm_membank_guassfilter_recon.png}
         \includegraphics[width=0.12\linewidth]{Figure_supp/MvTeccase/carpet/UniAD_heatmap.png}\hspace{-0.07cm}
         \includegraphics[width=0.12\linewidth]{Figure_supp/MvTeccase/carpet/ddpm_membank_guassfilter_heatmap.png}\hspace{-0.07cm}
         \includegraphics[width=0.12\linewidth]{Figure_supp/MvTeccase/carpet/UniAD_heatmapself.png}\hspace{-0.07cm}
         \includegraphics[width=0.12\linewidth]{Figure_supp/MvTeccase/carpet/ddpm_membank_guassfilter_heatmapself.png}
     \end{subfigure}
     \begin{subfigure}{\linewidth}
         \centering
         \includegraphics[width=0.12\linewidth]{Figure_supp/MvTeccase/grid/ori.png}\hspace{-0.07cm}
         \includegraphics[width=0.12\linewidth]{Figure_supp/MvTeccase/grid/gt.png}\hspace{-0.07cm}
         \includegraphics[width=0.12\linewidth]{Figure_supp/MvTeccase/grid/UniAD_recon.png}\hspace{-0.07cm}
         \includegraphics[width=0.12\linewidth]{Figure_supp/MvTeccase/grid/ddpm_membank_guassfilter_recon.png}
         \includegraphics[width=0.12\linewidth]{Figure_supp/MvTeccase/grid/UniAD_heatmap.png}\hspace{-0.07cm}
         \includegraphics[width=0.12\linewidth]{Figure_supp/MvTeccase/grid/ddpm_membank_guassfilter_heatmap.png}\hspace{-0.07cm}
         \includegraphics[width=0.12\linewidth]{Figure_supp/MvTeccase/grid/UniAD_heatmapself.png}\hspace{-0.07cm}
         \includegraphics[width=0.12\linewidth]{Figure_supp/MvTeccase/grid/ddpm_membank_guassfilter_heatmapself.png}
     \end{subfigure}
     \begin{subfigure}{\linewidth}
         \centering
         \includegraphics[width=0.12\linewidth]{Figure_supp/MvTeccase/metalnut/ori.png}\hspace{-0.07cm}
         \includegraphics[width=0.12\linewidth]{Figure_supp/MvTeccase/metalnut/gt.png}\hspace{-0.07cm}
         \includegraphics[width=0.12\linewidth]{Figure_supp/MvTeccase/metalnut/UniAD_recon.png}\hspace{-0.07cm}
         \includegraphics[width=0.12\linewidth]{Figure_supp/MvTeccase/metalnut/ddpm_membank_guassfilter_recon.png}
         \includegraphics[width=0.12\linewidth]{Figure_supp/MvTeccase/metalnut/UniAD_heatmap.png}\hspace{-0.07cm}
         \includegraphics[width=0.12\linewidth]{Figure_supp/MvTeccase/metalnut/ddpm_membank_guassfilter_heatmap.png}\hspace{-0.07cm}
         \includegraphics[width=0.12\linewidth]{Figure_supp/MvTeccase/metalnut/UniAD_heatmapself.png}\hspace{-0.07cm}
         \includegraphics[width=0.12\linewidth]{Figure_supp/MvTeccase/metalnut/ddpm_membank_guassfilter_heatmapself.png}
     \end{subfigure}
     \begin{subfigure}{\linewidth}
         \centering
         \includegraphics[width=0.12\linewidth]{Figure_supp/MvTeccase/screw/ori.png}\hspace{-0.07cm}
         \includegraphics[width=0.12\linewidth]{Figure_supp/MvTeccase/screw/gt.png}\hspace{-0.07cm}
         \includegraphics[width=0.12\linewidth]{Figure_supp/MvTeccase/screw/UniAD_recon.png}\hspace{-0.07cm}
         \includegraphics[width=0.12\linewidth]{Figure_supp/MvTeccase/screw/ddpm_membank_guassfilter_recon.png}
         \includegraphics[width=0.12\linewidth]{Figure_supp/MvTeccase/screw/UniAD_heatmap.png}\hspace{-0.07cm}
         \includegraphics[width=0.12\linewidth]{Figure_supp/MvTeccase/screw/ddpm_membank_guassfilter_heatmap.png}\hspace{-0.07cm}
         \includegraphics[width=0.12\linewidth]{Figure_supp/MvTeccase/screw/UniAD_heatmapself.png}\hspace{-0.07cm}
         \includegraphics[width=0.12\linewidth]{Figure_supp/MvTeccase/screw/ddpm_membank_guassfilter_heatmapself.png}
     \end{subfigure}
     \begin{subfigure}{\linewidth}
         \centering
         \includegraphics[width=0.12\linewidth]{Figure_supp/MvTeccase/tile/ori.png}\hspace{-0.07cm}
         \includegraphics[width=0.12\linewidth]{Figure_supp/MvTeccase/tile/gt.png}\hspace{-0.07cm}
         \includegraphics[width=0.12\linewidth]{Figure_supp/MvTeccase/tile/UniAD_recon.png}\hspace{-0.07cm}
         \includegraphics[width=0.12\linewidth]{Figure_supp/MvTeccase/tile/ddpm_membank_guassfilter_recon.png}
         \includegraphics[width=0.12\linewidth]{Figure_supp/MvTeccase/tile/UniAD_heatmap.png}\hspace{-0.07cm}
         \includegraphics[width=0.12\linewidth]{Figure_supp/MvTeccase/tile/ddpm_membank_guassfilter_heatmap.png}\hspace{-0.07cm}
         \includegraphics[width=0.12\linewidth]{Figure_supp/MvTeccase/tile/UniAD_heatmapself.png}\hspace{-0.07cm}
         \includegraphics[width=0.12\linewidth]{Figure_supp/MvTeccase/tile/ddpm_membank_guassfilter_heatmapself.png}
     \end{subfigure}
     \begin{subfigure}{\linewidth}
         \centering
         \includegraphics[width=0.12\linewidth]{Figure_supp/MvTeccase/transistor/ori.png}\hspace{-0.07cm}
         \includegraphics[width=0.12\linewidth]{Figure_supp/MvTeccase/transistor/gt.png}\hspace{-0.07cm}
         \includegraphics[width=0.12\linewidth]{Figure_supp/MvTeccase/transistor/UniAD_recon.png}\hspace{-0.07cm}
         \includegraphics[width=0.12\linewidth]{Figure_supp/MvTeccase/transistor/ddpm_membank_guassfilter_recon.png}
         \includegraphics[width=0.12\linewidth]{Figure_supp/MvTeccase/transistor/UniAD_heatmap.png}\hspace{-0.07cm}
         \includegraphics[width=0.12\linewidth]{Figure_supp/MvTeccase/transistor/ddpm_membank_guassfilter_heatmap.png}\hspace{-0.07cm}
         \includegraphics[width=0.12\linewidth]{Figure_supp/MvTeccase/transistor/UniAD_heatmapself.png}\hspace{-0.07cm}
         \includegraphics[width=0.12\linewidth]{Figure_supp/MvTeccase/transistor/ddpm_membank_guassfilter_heatmapself.png}
     \end{subfigure}
     \hfill
     \begin{subfigure}{0.12\linewidth}
         \centering
         \includegraphics[width=\linewidth]{Figure_supp/MvTeccase/Zipper/ori.png}
         \caption*{Input}
     \end{subfigure}\hspace{-0.07cm}
     \begin{subfigure}{0.12\linewidth}
         \centering
         \includegraphics[width=\linewidth]{Figure_supp/MvTeccase/Zipper/gt.png}
         \caption*{GT}
     \end{subfigure}\hspace{-0.07cm}
     \begin{subfigure}{0.12\linewidth}
         \centering
         \includegraphics[width=\linewidth]{Figure_supp/MvTeccase/Zipper/UniAD_recon.png}
         \caption*{UniAD}
     \end{subfigure}\hspace{-0.07cm}
     \begin{subfigure}{0.12\linewidth}
         \centering
         \includegraphics[width=\linewidth]{Figure_supp/MvTeccase/Zipper/ddpm_membank_guassfilter_recon.png}
         \caption*{Ours}
     \end{subfigure}
     \begin{subfigure}{0.12\linewidth}
         \centering
         \includegraphics[width=\linewidth]{Figure_supp/MvTeccase/Zipper/UniAD_heatmap.png}
         \caption*{UniAD}
     \end{subfigure}\hspace{-0.07cm}
     \begin{subfigure}{0.12\linewidth}
         \centering
         \includegraphics[width=\linewidth]{Figure_supp/MvTeccase/Zipper/ddpm_membank_guassfilter_heatmap.png}
         \caption*{Ours}
     \end{subfigure}\hspace{-0.07cm}
     \begin{subfigure}{0.12\linewidth}
         \centering
         \includegraphics[width=\linewidth]{Figure_supp/MvTeccase/Zipper/UniAD_heatmapself.png}
         \caption*{UniAD}
     \end{subfigure}\hspace{-0.07cm}
     \begin{subfigure}{0.12\linewidth}
         \centering
         \includegraphics[width=\linewidth]{Figure_supp/MvTeccase/Zipper/ddpm_membank_guassfilter_heatmapself.png}
         \caption*{Ours}
     \end{subfigure}
    \caption{Visualization of reconstruction and localization results on MVTec-AD \cite{MVTec}. The first two columns show the original image and the anomalies ground truth, the third and fourth columns show the visual reconstruction of UniAD and our method, the fifth and sixth columns show the anomaly localization normalized on single category for UniAD and our method, and the seventh and eighth columns show the anomaly localization normalized on single image for UniAD and our method.
    }
    \label{fig:visualization_MvTec}
\end{figure*}

\begin{figure*}[t]
     \centering
     \begin{subfigure}{\linewidth}
         \centering
         \includegraphics[width=0.12\linewidth]{Figure_supp/MPDDcase/bracket_black2/ori.png}\hspace{-0.07cm}
         \includegraphics[width=0.12\linewidth]{Figure_supp/MPDDcase/bracket_black2/gt.png}\hspace{-0.07cm}
         \includegraphics[width=0.12\linewidth]{Figure_supp/MPDDcase/bracket_black2/UniAD_recon.png}\hspace{-0.07cm}
         \includegraphics[width=0.12\linewidth]{Figure_supp/MPDDcase/bracket_black2/ddpm_recon.png}
         \includegraphics[width=0.12\linewidth]{Figure_supp/MPDDcase/bracket_black2/UniAD_heatmap.png}\hspace{-0.07cm}
         \includegraphics[width=0.12\linewidth]{Figure_supp/MPDDcase/bracket_black2/ddpm_heatmap.png}\hspace{-0.07cm}
         \includegraphics[width=0.12\linewidth]{Figure_supp/MPDDcase/bracket_black2/UniAD_heatmapself.png}\hspace{-0.07cm}
         \includegraphics[width=0.12\linewidth]{Figure_supp/MPDDcase/bracket_black2/ddpm_heatmapself.png}
     \end{subfigure}
     \begin{subfigure}{\linewidth}
         \centering
         \includegraphics[width=0.12\linewidth]{Figure_supp/MPDDcase/bracket_black1/ori.png}\hspace{-0.07cm}
         \includegraphics[width=0.12\linewidth]{Figure_supp/MPDDcase/bracket_black1/gt.png}\hspace{-0.07cm}
         \includegraphics[width=0.12\linewidth]{Figure_supp/MPDDcase/bracket_black1/UniAD_recon.png}\hspace{-0.07cm}
         \includegraphics[width=0.12\linewidth]{Figure_supp/MPDDcase/bracket_black1/ddpm_recon.png}
         \includegraphics[width=0.12\linewidth]{Figure_supp/MPDDcase/bracket_black1/UniAD_heatmap.png}\hspace{-0.07cm}
         \includegraphics[width=0.12\linewidth]{Figure_supp/MPDDcase/bracket_black1/ddpm_heatmap.png}\hspace{-0.07cm}
         \includegraphics[width=0.12\linewidth]{Figure_supp/MPDDcase/bracket_black1/UniAD_heatmapself.png}\hspace{-0.07cm}
         \includegraphics[width=0.12\linewidth]{Figure_supp/MPDDcase/bracket_black1/ddpm_heatmapself.png}
     \end{subfigure}
     \begin{subfigure}{\linewidth}
         \centering
         \includegraphics[width=0.12\linewidth]{Figure_supp/MPDDcase/bracket_white1/ori.png}\hspace{-0.07cm}
         \includegraphics[width=0.12\linewidth]{Figure_supp/MPDDcase/bracket_white1/gt.png}\hspace{-0.07cm}
         \includegraphics[width=0.12\linewidth]{Figure_supp/MPDDcase/bracket_white1/UniAD_recon.png}\hspace{-0.07cm}
         \includegraphics[width=0.12\linewidth]{Figure_supp/MPDDcase/bracket_white1/ddpm_recon.png}
         \includegraphics[width=0.12\linewidth]{Figure_supp/MPDDcase/bracket_white1/UniAD_heatmap.png}\hspace{-0.07cm}
         \includegraphics[width=0.12\linewidth]{Figure_supp/MPDDcase/bracket_white1/ddpm_heatmap.png}\hspace{-0.07cm}
         \includegraphics[width=0.12\linewidth]{Figure_supp/MPDDcase/bracket_white1/UniAD_heatmapself.png}\hspace{-0.07cm}
         \includegraphics[width=0.12\linewidth]{Figure_supp/MPDDcase/bracket_white1/ddpm_heatmapself.png}
     \end{subfigure}
     \begin{subfigure}{\linewidth}
         \centering
         \includegraphics[width=0.12\linewidth]{Figure_supp/MPDDcase/connector1/ori.png}\hspace{-0.07cm}
         \includegraphics[width=0.12\linewidth]{Figure_supp/MPDDcase/connector1/gt.png}\hspace{-0.07cm}
         \includegraphics[width=0.12\linewidth]{Figure_supp/MPDDcase/connector1/UniAD_recon.png}\hspace{-0.07cm}
         \includegraphics[width=0.12\linewidth]{Figure_supp/MPDDcase/connector1/ddpm_recon.png}
         \includegraphics[width=0.12\linewidth]{Figure_supp/MPDDcase/connector1/UniAD_heatmap.png}\hspace{-0.07cm}
         \includegraphics[width=0.12\linewidth]{Figure_supp/MPDDcase/connector1/ddpm_heatmap.png}\hspace{-0.07cm}
         \includegraphics[width=0.12\linewidth]{Figure_supp/MPDDcase/connector1/UniAD_heatmapself.png}\hspace{-0.07cm}
         \includegraphics[width=0.12\linewidth]{Figure_supp/MPDDcase/connector1/ddpm_heatmapself.png}
     \end{subfigure}
     \begin{subfigure}{\linewidth}
         \centering
         \includegraphics[width=0.12\linewidth]{Figure_supp/MPDDcase/metal1/ori.png}\hspace{-0.07cm}
         \includegraphics[width=0.12\linewidth]{Figure_supp/MPDDcase/metal1/gt.png}\hspace{-0.07cm}
         \includegraphics[width=0.12\linewidth]{Figure_supp/MPDDcase/metal1/UniAD_recon.png}\hspace{-0.07cm}
         \includegraphics[width=0.12\linewidth]{Figure_supp/MPDDcase/metal1/ddpm_recon.png}
         \includegraphics[width=0.12\linewidth]{Figure_supp/MPDDcase/metal1/UniAD_heatmap.png}\hspace{-0.07cm}
         \includegraphics[width=0.12\linewidth]{Figure_supp/MPDDcase/metal1/ddpm_heatmap.png}\hspace{-0.07cm}
         \includegraphics[width=0.12\linewidth]{Figure_supp/MPDDcase/metal1/UniAD_heatmapself.png}\hspace{-0.07cm}
         \includegraphics[width=0.12\linewidth]{Figure_supp/MPDDcase/metal1/ddpm_heatmapself.png}
     \end{subfigure}
     \begin{subfigure}{\linewidth}
         \centering
         \includegraphics[width=0.12\linewidth]{Figure_supp/MPDDcase/metal2/ori.png}\hspace{-0.07cm}
         \includegraphics[width=0.12\linewidth]{Figure_supp/MPDDcase/metal2/gt.png}\hspace{-0.07cm}
         \includegraphics[width=0.12\linewidth]{Figure_supp/MPDDcase/metal2/UniAD_recon.png}\hspace{-0.07cm}
         \includegraphics[width=0.12\linewidth]{Figure_supp/MPDDcase/metal2/ddpm_recon.png}
         \includegraphics[width=0.12\linewidth]{Figure_supp/MPDDcase/metal2/UniAD_heatmap.png}\hspace{-0.07cm}
         \includegraphics[width=0.12\linewidth]{Figure_supp/MPDDcase/metal2/ddpm_heatmap.png}\hspace{-0.07cm}
         \includegraphics[width=0.12\linewidth]{Figure_supp/MPDDcase/metal2/UniAD_heatmapself.png}\hspace{-0.07cm}
         \includegraphics[width=0.12\linewidth]{Figure_supp/MPDDcase/metal2/ddpm_heatmapself.png}
     \end{subfigure}
     \begin{subfigure}{\linewidth}
         \centering
         \includegraphics[width=0.12\linewidth]{Figure_supp/MPDDcase/tube1/ori.png}\hspace{-0.07cm}
         \includegraphics[width=0.12\linewidth]{Figure_supp/MPDDcase/tube1/gt.png}\hspace{-0.07cm}
         \includegraphics[width=0.12\linewidth]{Figure_supp/MPDDcase/tube1/UniAD_recon.png}\hspace{-0.07cm}
         \includegraphics[width=0.12\linewidth]{Figure_supp/MPDDcase/tube1/ddpm_recon.png}
         \includegraphics[width=0.12\linewidth]{Figure_supp/MPDDcase/tube1/UniAD_heatmap.png}\hspace{-0.07cm}
         \includegraphics[width=0.12\linewidth]{Figure_supp/MPDDcase/tube1/ddpm_heatmap.png}\hspace{-0.07cm}
         \includegraphics[width=0.12\linewidth]{Figure_supp/MPDDcase/tube1/UniAD_heatmapself.png}\hspace{-0.07cm}
         \includegraphics[width=0.12\linewidth]{Figure_supp/MPDDcase/tube1/ddpm_heatmapself.png}
     \end{subfigure}
     \begin{subfigure}{\linewidth}
         \centering
         \includegraphics[width=0.12\linewidth]{Figure_supp/MPDDcase/tube2/ori.png}\hspace{-0.07cm}
         \includegraphics[width=0.12\linewidth]{Figure_supp/MPDDcase/tube2/gt.png}\hspace{-0.07cm}
         \includegraphics[width=0.12\linewidth]{Figure_supp/MPDDcase/tube2/UniAD_recon.png}\hspace{-0.07cm}
         \includegraphics[width=0.12\linewidth]{Figure_supp/MPDDcase/tube2/ddpm_recon.png}
         \includegraphics[width=0.12\linewidth]{Figure_supp/MPDDcase/tube2/UniAD_heatmap.png}\hspace{-0.07cm}
         \includegraphics[width=0.12\linewidth]{Figure_supp/MPDDcase/tube2/ddpm_heatmap.png}\hspace{-0.07cm}
         \includegraphics[width=0.12\linewidth]{Figure_supp/MPDDcase/tube2/UniAD_heatmapself.png}\hspace{-0.07cm}
         \includegraphics[width=0.12\linewidth]{Figure_supp/MPDDcase/tube2/ddpm_heatmapself.png}
     \end{subfigure}
     \hfill
     \begin{subfigure}{0.12\linewidth}
         \centering
         \includegraphics[width=\linewidth]{Figure_supp/MPDDcase/tube3/ori.png}
         \caption*{Input}
     \end{subfigure}\hspace{-0.07cm}
     \begin{subfigure}{0.12\linewidth}
         \centering
         \includegraphics[width=\linewidth]{Figure_supp/MPDDcase/tube3/gt.png}
         \caption*{GT}
     \end{subfigure}\hspace{-0.07cm}
     \begin{subfigure}{0.12\linewidth}
         \centering
         \includegraphics[width=\linewidth]{Figure_supp/MPDDcase/tube3/UniAD_recon.png}
         \caption*{UniAD}
     \end{subfigure}\hspace{-0.07cm}
     \begin{subfigure}{0.12\linewidth}
         \centering
         \includegraphics[width=\linewidth]{Figure_supp/MPDDcase/tube3/ddpm_recon.png}
         \caption*{Ours}
     \end{subfigure}
     \begin{subfigure}{0.12\linewidth}
         \centering
         \includegraphics[width=\linewidth]{Figure_supp/MPDDcase/tube3/UniAD_heatmap.png}
         \caption*{UniAD}
     \end{subfigure}\hspace{-0.07cm}
     \begin{subfigure}{0.12\linewidth}
         \centering
         \includegraphics[width=\linewidth]{Figure_supp/MPDDcase/tube3/ddpm_heatmap.png}
         \caption*{Ours}
     \end{subfigure}\hspace{-0.07cm}
     \begin{subfigure}{0.12\linewidth}
         \centering
         \includegraphics[width=\linewidth]{Figure_supp/MPDDcase/tube3/UniAD_heatmapself.png}
         \caption*{UniAD}
     \end{subfigure}\hspace{-0.07cm}
     \begin{subfigure}{0.12\linewidth}
         \centering
         \includegraphics[width=\linewidth]{Figure_supp/MPDDcase/tube3/ddpm_heatmapself.png}
         \caption*{Ours}
     \end{subfigure}
    \caption{Visualization of reconstruction and localization results on MPDD\cite{MPDD}. The arrangement of the images is the same as in Figure \ref{fig:visualization_MvTec}.  
    }
    \label{fig:visualization_MPDD}
\end{figure*}

{\small
\bibliographystyle{ieee_fullname}
\bibliography{reference}
}